\documentclass[a4paper,fleqn]{cas-dc}

\usepackage[numbers]{natbib}

\def\tsc#1{\csdef{#1}{\textsc{\lowercase{#1}}\xspace}}
\tsc{WGM}
\tsc{QE}
\tsc{EP}
\tsc{PMS}
\tsc{BEC}
\tsc{DE}
\newcommand{\tableref}[1]{\hyperref[#1]{\textcolor{blue}{Table }\textup{(\ref{#1})}}}
\newcommand{\figref}[1]{\hyperref[#1]{\textcolor{blue}{Fig.}\textup{(\ref{#1})}}}
\newcommand{\equref}[1]{\hyperref[#1]{\textcolor{blue}{Eq.}\textup{\ref{#1}}}}
\usepackage{algpseudocode}
\usepackage{amsmath}
\usepackage{graphicx}
\usepackage[misc]{ifsym}
\usepackage{utfsym}
\usepackage{epstopdf}
\usepackage{subfigure}
\usepackage{algorithm}
\usepackage{tabularx}
\usepackage{soul}
\usepackage{xcolor}

\begin{document}
\let\WriteBookmarks\relax
\def\floatpagepagefraction{1}
\def\textpagefraction{.001}
\shortauthors{Wenjie~Luo et~al.}

\title [mode = title]{MFC-RFNet: A Multi-scale Guided Rectified Flow Network for Radar Sequence Prediction}                      
\tnotemark[1]

\tnotetext[1]{This work was supported in part by the Science and Technology Department of Sichuan Province under Grant 2024ZYD0089; and in part by Yibin University Science and Technology under Grant 2024XJYY03, Grant 2023YY02 and Grant 2023YY05. (\textit{WenJie Luo and Chuanhu Deng contributed equally to this work.}) (\textit{Corresponding author: C.R. Li})}

\affiliation[1]{organization={Yibin University},
            addressline={School of Computer Science and Technology (School of Artificial Intelligence)}, 
            city={Yibin},
            postcode={644000}, 
            country={China}}
\affiliation[2]{organization={Chongqing University of Technology},
            addressline={College of Computer Science and Engineering}, 
            city={Chongqing},
            postcode={404100}, 
            country={China}}
\affiliation[3]{organization={Yibin Meteorological Bureau},
            city={Yibin},
            postcode={644000}, 
            country={China}}

\author[1,2]{Wenjie~Luo}
\ead{lwj018@stu.cqut.edu.cn}
\author[1]{Chaorong~Li}
\cormark[1]
\ead{lichaorong88@163.com}

\author[1,2]{Chuanhu~Deng}
\ead{onlyiyou@stu.cqut.edu.cn}

\author[3]{Rongyao~Deng}
\ead{woshidengry@163.com}

\author[1]{Qiang~Yang}
\ead{604741717@qq.com}

\cortext[cor1]{Corresponding author}

\begin{abstract}
Accurate and high-resolution precipitation nowcasting from radar echo sequences is crucial for disaster mitigation and economic planning, yet it remains a significant challenge. Key difficulties include modeling complex multi-scale evolution, correcting inter-frame feature misalignment caused by displacement, and efficiently capturing long-range spatiotemporal context without sacrificing spatial fidelity. To address these issues, we present the Multi-scale Feature Communication Rectified Flow (RF) Network (MFC-RFNet), a generative framework that integrates multi-scale communication with guided feature fusion. To enhance multi-scale fusion while retaining fine detail, a Wavelet-Guided Skip Connection (WGSC) preserves high-frequency components, and a Feature Communication Module (FCM) promotes bidirectional cross-scale interaction. To correct inter-frame displacement, a Condition-Guided Spatial Transform Fusion (CGSTF) learns spatial transforms from conditioning echoes to align shallow features. The backbone adopts rectified flow training to learn near-linear probability-flow trajectories, enabling few-step sampling with stable fidelity. Additionally, lightweight 
Vision-RWKV (RWKV) blocks are placed at the encoder tail, the bottleneck, and the first decoder layer to capture long-range spatiotemporal dependencies at low spatial resolutions with moderate compute. Evaluations on four public datasets (SEVIR, MeteoNet, Shanghai, and CIKM) demonstrate consistent improvements over strong baselines, yielding clearer echo morphology at higher rain-rate thresholds and sustained skill at longer lead times. These results suggest that the proposed synergy of RF training with scale-aware communication, spatial alignment, and frequency-aware fusion presents an effective and robust approach for radar-based nowcasting.
\end{abstract}

\begin{keywords}
Precipitation Nowcasting\sep Multi-scale Feature Fusion\sep Rectified Flow\sep Generative Models\sep Spatiotemporal Forecasting
\end{keywords}

\maketitle

\section{Introduction}
Radar sequence prediction (RSP)—forecasting future radar echo fields from recent sequences—underpins operational precipitation nowcasting at 0–2 h lead times for flood early warning, urban traffic management, and precision agriculture \cite{wmo2017}. Within these short horizons, models must capture large-scale advection while tracking the rapid growth, decay, rotation, and deformation of localized convection under tight latency constraints. The radar echo field is intermittent, nonlinear, and strongly multiscale, making minute-scale evolution difficult to model without sacrificing spatial detail or intensity fidelity \cite{sun2014}. These demands naturally motivate end-to-end learning on radar sequences and place RSP at the center of our formulation; accordingly, we next examine learning-based RSP architectures—from recurrent cells to attention and modern generative/flow formulations—that have driven recent progress.\par
\begin{figure}[t]
  \centering{\includegraphics[scale=0.23]{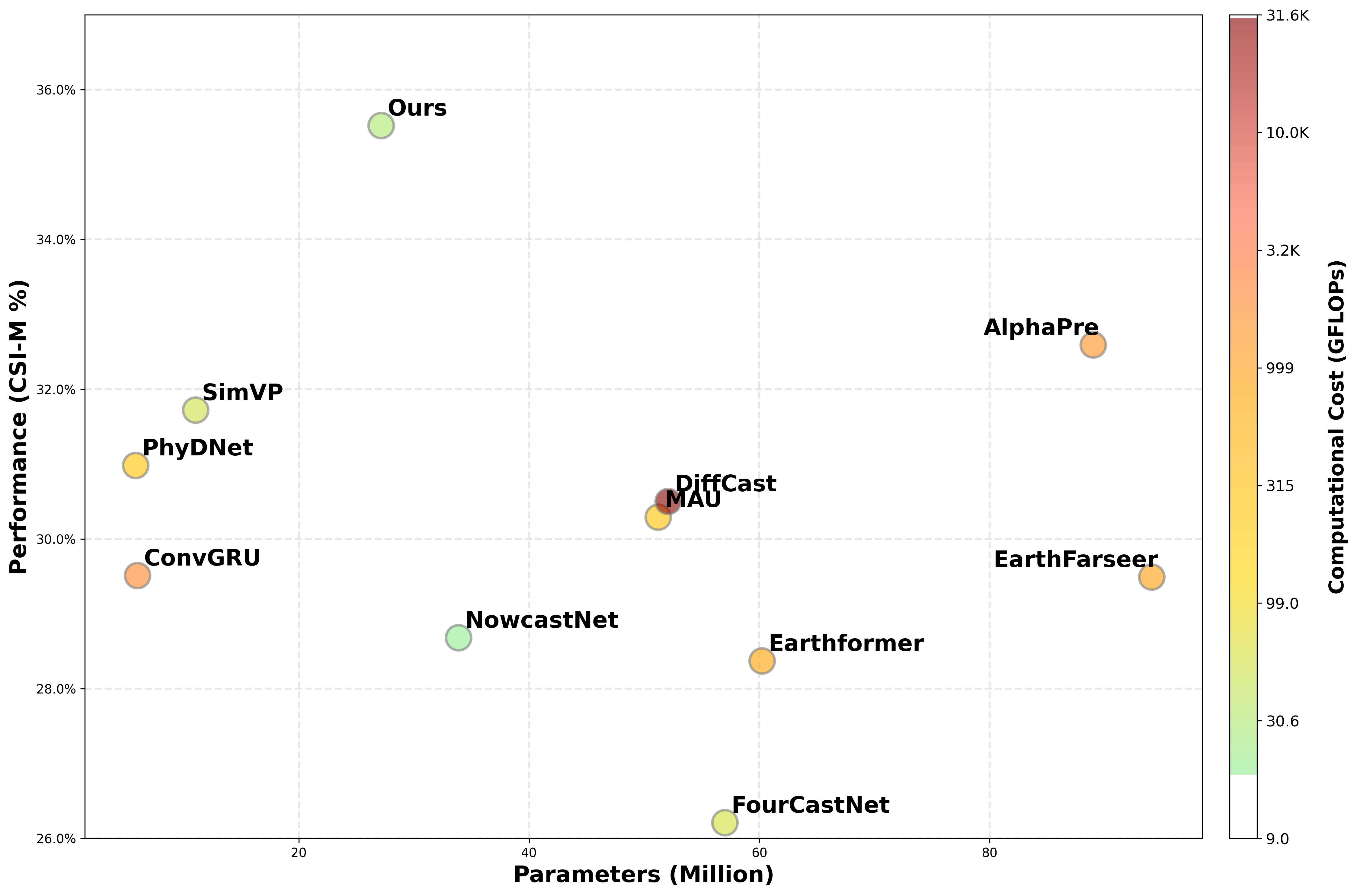}}
\caption{\textbf{Pareto plot of parameters, compute, and performance on SEVIR.}
  Vertical axis: CSI-M (\%); horizontal axis: number of parameters (Million).
  The color bar encodes computational cost in GFLOPs—darker indicates higher cost.}
  \label{fig:sevir_pareto}
\end{figure}
Deep learning-based radar sequence models have advanced nowcasting substantially. ConvLSTM \cite{shi2015convlstm} first demonstrated clear gains over classical extrapolation on real radar data, and TrajGRU \cite{shi2017trajgru} improved the representation of local dynamics such as rotation and deformation through location-variant recurrent connections. To enlarge receptive fields and capture long-range dependencies, attention-based and large-context models (MetNet \cite{metnet2020} and MetNet-2 \cite{metnet22022}) achieved strong probabilistic forecasts in the short to medium range. In parallel, generative paradigms—from GANs to diffusion—have improved uncertainty quantification and detail quality for precipitation fields: DGMR \cite{ravuri2021nature} reported strong skill on heavy rainfall, while latent and residual diffusion variants further alleviated high-value underestimation and over-smoothing \cite{prediff2023,liu2024residual}. More recently, Flow Matching \cite{lipman2022flow} and Rectified Flow \cite{liu2022rectified,reflow2024} have emerged as probability-flow frameworks that parameterize generation along near-straight trajectories in distribution space, offering a favorable efficiency–quality trade-off with few sampling steps and stable training.

While these trends confirm that deep learning is a promising route for nowcasting, several critical challenges persist that prevent current models from reaching their full potential. A primary issue is the insufficient coordination between multi-scale features, where the lack of effective communication between shallow, detail-rich layers and deep, context-aware layers can lead to blurry or structurally inconsistent predictions \cite{he2024long,zhang2023skilful}. Furthermore, the inherent motion of precipitation systems induces inter-frame displacement, causing spatial misalignment of features at early processing stages that degrades performance if not explicitly corrected \cite{Wu2021_MotionRNN,LeGuen2020_PhyDNet}. Compounding these issues is the high computational cost of modeling long-range spatiotemporal dependencies for large-scale events \cite{gao2022earthformer,pathak2022fourcastnet}. Addressing this trio of challenges in a unified framework is the central motivation for the architecture developed in this work.

\begin{figure*}[htbp]
  \centering{\includegraphics[scale=0.52]{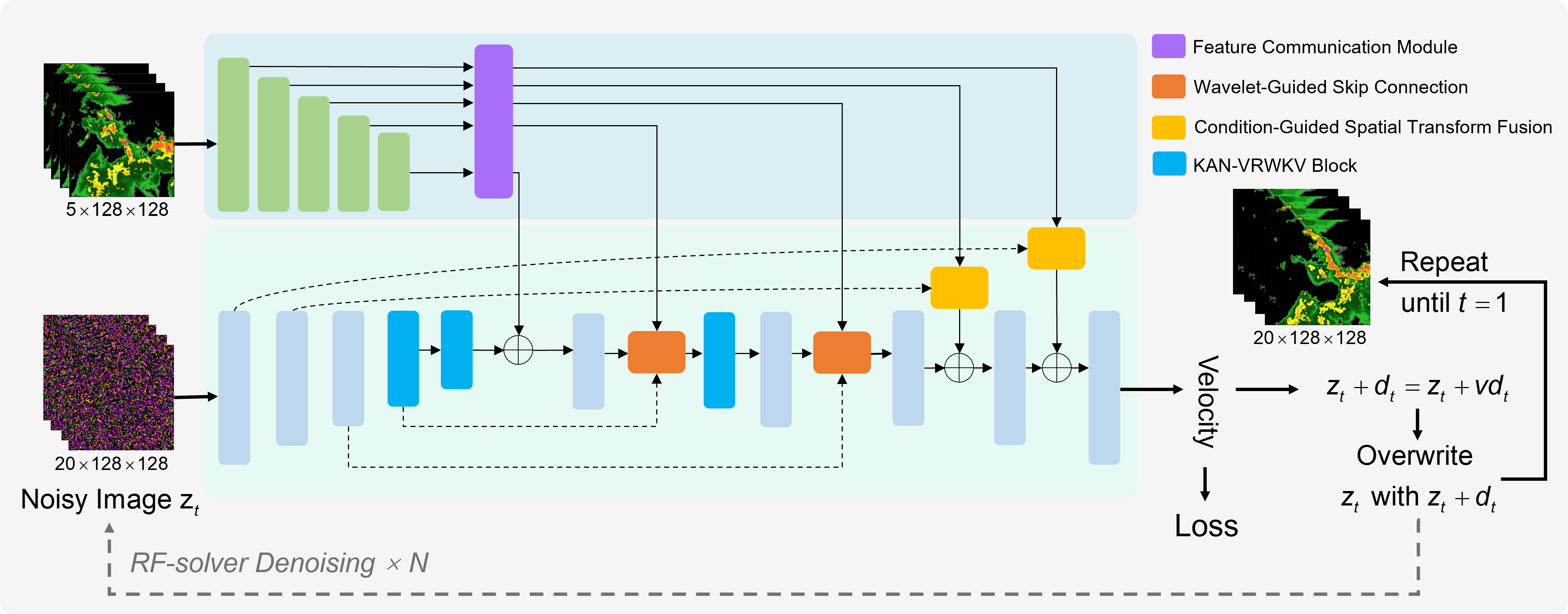}}
\caption{\textbf{MFC-RFNet overall architecture.} Top: the conditional encoder extracts multi-scale features; FCM performs cross-scale communication; CGSTF aligns shallow features; WGSC modulates skip fusion. Bottom: the RF generator updates $z_t$ with few ODE steps ($z_{t+\Delta t} = z_t + v_{\theta}\,\Delta t$) to produce $K$ future frames, with VRWKV placed in deep layers to supply long-range context. }
  \label{fig:fig1}
\end{figure*}
To address these distinct challenges, we propose the Multi-scale Feature Communication Rectified Flow Network (MFC-RFNet), a generative architecture tailored for radar-based precipitation nowcasting. MFC-RFNet introduces a set of complementary modules, each targeting a specific bottleneck: (i) to improve cross-scale coordination, a Feature Communication Module (FCM) establishes bidirectional pathways across scales and enables pixel-wise cross-scale selection, improving the coordination between global context and local detail; (ii) to mitigate shallow-layer spatial misalignment, a Condition-Guided Spatial Transform Fusion (CGSTF) estimates a displacement field from conditioning echoes and performs differentiable alignment of shallow backbone features before fusion, which is related to Spatial Transformer Networks but conditioned on the input radar sequence \cite{jaderberg2015stn}; (iii) to further refine feature fusion with a focus on fidelity, a Wavelet-Guided Skip Connection (WGSC) applies discrete wavelet decomposition to the conditional features and learns frequency-aware gates to modulate encoder-decoder skip pathways, maintaining global structural consistency while preserving high-frequency components \cite{waveletunet2018}; and (iv) to efficiently model long-range dependencies, a compact Vision-RWKV (RWKV) block is placed at the deepest stages to supply global spatiotemporal context with moderate compute \cite{rwkv2023,vrwkv2024}. These components are integrated within a RF-based generative backbone; training follows the RF paradigm, which parameterizes the generative process as a near-linear probability flow and supports few-step sampling while preserving fidelity. Beyond module-level design, we highlight the accuracy–efficiency trade-off: on SEVIR, MFC-RFNet exhibits a competitive balance among parameter count, computational cost, and forecasting performance. As shown in Fig.~\ref{fig:sevir_pareto}, our model tends to lie near the Pareto frontier compared with recent baselines. Architectural specifics and ablations are detailed in Section~\ref{sec:Methodology}.\par

In summary, our main contributions are as follows:
\begin{itemize}
    \item We present MFC-RFNet, a unified generative framework that combines RF training with scale-aware communication, shallow-layer spatial alignment, 
    \\frequency-aware skip fusion, and efficient deep-stage long-sequence modeling in a radar-only setting.
    \item We design three guided modules for the conditional pathway: an FCM enabling bidirectional information flow across scales; a CGSTF that performs explicit alignment of shallow features via learned spatial transforms from conditioning echoes; and a WGSC that controls skip fusion using wavelet-based frequency cues.
    \item Experiments on four public benchmarks (SEVIR, MeteoNet, Shanghai, and CIKM) show consistent improvements over strong baselines across rain-rate thresholds and lead times, supporting the effectiveness and generalization of the proposed design.
\end{itemize}

\section{Related Work}

\subsection{Deterministic and Generative Paradigms in Nowcasting}
Deterministic models learn a direct mapping from a history of radar echoes to a single future sequence. Let $\mathbf{c}=\mathbf{X}_{t-J+1:t}$ denote the conditioning history and $\mathbf{z}_0=\mathbf{X}_{t+1:t+K}$ the future tensor. A predictor $f_\theta$ outputs $\widehat{\mathbf{z}}=f_\theta(\mathbf{c})$ by minimizing an empirical loss such as MSE or $\ell_1$. Architectures in this category include recurrent models \cite{wang2019memory,luo2020pfst}, U-Net–based convolutional predictors \cite{turzi2025ssa,wang2025rainhcnet}, and Transformer-based large-context designs \cite{yang2022aa,bai2022rainformer}. While effective at capturing general advection, pointwise objectives often average over multiple plausible futures, which tends to blur fine structures and underestimate high intensities over longer lead times. Rollout accumulation can further degrade skill when predictions are fed back as inputs.

Generative models instead aim to approximate the full conditional distribution $p_\theta(\mathbf{z}\mid\mathbf{c})$ and sample ensembles $\mathbf{z}\sim p_\theta(\cdot\mid\mathbf{c})$ to represent forecast uncertainty. Early successes with adversarial training produced sharper realizations with improved heavy-rain skill \cite{ling2024tu2net}. More recent diffusion and probability-flow frameworks learn denoising scores or velocity fields, enabling high-fidelity synthesis \cite{li2024precipitation,li2025extreme}; in particular, rectified flow formulations enable few-step deterministic sampling with stable optimization \cite{liu2023instaflow}. These approaches improve uncertainty quantification and help preserve high-frequency details. However, their predictive skill depends critically on the quality of the conditional representation: weaknesses in the encoder, such as poor cross-scale coordination or uncorrected shallow-stage misalignment, create a bottleneck that limits the potential of even the most powerful generative backbone. This dependency presents a clear opportunity to advance the field by designing specialized conditional architectures that strengthen multi-scale communication, perform early alignment, and regulate feature fusion—precisely the challenges addressed in this work.

\subsection{Rectified Flow Models}
Rectified Flow (RF) advances conditional generation by learning a deterministic probability-flow through an ordinary differential equation (ODE). Let $Z_t$ denote the latent (or data) variable at time $t\!\in\![0,1]$ and $v_\theta$ a learnable velocity field. RF defines a near-linear path from noise to data via
\begin{equation}
  dZ_t = v_\theta(Z_t,t)\,dt,\qquad t\in[0,1],
\end{equation}
and optimizes the velocity to match straight interpolants between samples:
\begin{equation}
\begin{aligned}
  \min_{\theta}\;&\int_0^1 \mathbb{E}\!\left[\,\|X_1 - X_0 - v_\theta(X_t,t)\|^2\,\right] dt,\\
  &X_t = tX_1 + (1-t)X_0 .
\end{aligned}
\end{equation}

Compared with stochastic diffusion processes, RF offers (i) computational efficiency-deterministic ODE sampling typically requires far fewer steps; (ii) training stability-near-linear transport mitigates error accumulation associated with curved trajectories; and (iii) generation quality-recent image/video applications report competitive fidelity with concise samplers. Despite these merits, the use of RF in unified spatiotemporal forecasting of radar sequences is still comparatively limited, especially when it comes to conditional encoders that must coordinate information across scales and align shallow features before decoding. This motivates architectures that pair RF backbones with scale-aware communication and feature-fusion mechanisms tailored to radar-based nowcasting.

\subsection{Architectural Innovations for Spatiotemporal Representation}
Effective spatiotemporal forecasting relies on architectural components that address multiscale fusion, spatial alignment, frequency-aware modeling, and long-range temporal dependencies.

\paragraph{Multiscale feature fusion}
Multiscale representation is fundamental in vision and geoscience \cite{Bi2023_PanguWeather}. Hierarchical designs such as CFPN \cite{du2025cross} and IMFA \cite{zhang2023towards} introduce top-down and bottom-up pathways to combine semantic and spatial information. In forecasting models, however, fusion is often implemented via concatenation or summation, which can underuse complementary cues across resolutions and impede the transfer of global context to fine-grained localization. Recent work explores bidirectional pathways, lateral connections, and cross-scale selection to enhance coordination, motivating cross-scale communication modules that operate at multiple pyramid levels \cite{li2024multi}.

\paragraph{Spatial transforms and shallow-layer alignment}
Non-rigid displacements are central to spatial localization. Generic tools such as spatial transformer networks \cite{jaderberg2015stn} and deformable convolutions \cite{wang2023internimage} enable learnable warping, while motion-centric sequence models (e.g., MotionRNN, PhyDNet) introduce motion-aware priors \cite{Wu2021_MotionRNN,LeGuen2020_PhyDNet}. For radar sequences, local deformations (e.g., splitting/merging) often arise in early layers and can propagate errors if not corrected, motivating shallow-layer alignment that estimates displacement fields from the conditioning inputs and aligns features before deeper aggregation.

\paragraph{Frequency-domain and wavelet-based modeling}
Frequency analysis separates structure from detail and can regularize fusion. Global filtering families (e.g., SpectFormer/GFNet) expand receptive fields via learned spectral operators \cite{patro2025spectformer,rao2023gfnet}, and wavelet-guided architectures—including Transformers for denoising and diffusion models for generation—leverage multiresolution analysis via DWT/IWT for efficient reconstruction and fast sampling \cite{li2024ewt,phung2023wavelet}. Wavelet methods also appear in scale-aware evaluation for meteorology \cite{taie2024statistical}. Injecting frequency cues into skip connections helps preserve high-frequency components while maintaining large-scale consistency, suggesting wavelet-guided gating as a practical control mechanism during fusion.

\paragraph{Efficient long-sequence modeling}
As forecast horizons and spatial resolutions increase, quadratic self-attention becomes computationally prohibitive. Efficient alternatives include I/O-aware exact attention \cite{dao2023flashattention} and linear-time state-space models such as Mamba and RWKV (with a vision adaptation, VRWKV) \cite{gu2023mamba,vrwkv2024}. A common strategy is to place these compact modules at deep stages (e.g., the bottleneck) to inject global spatiotemporal context where the receptive field is largest.


\begin{figure*}[htbp]
  \centering{\includegraphics[scale=0.34]{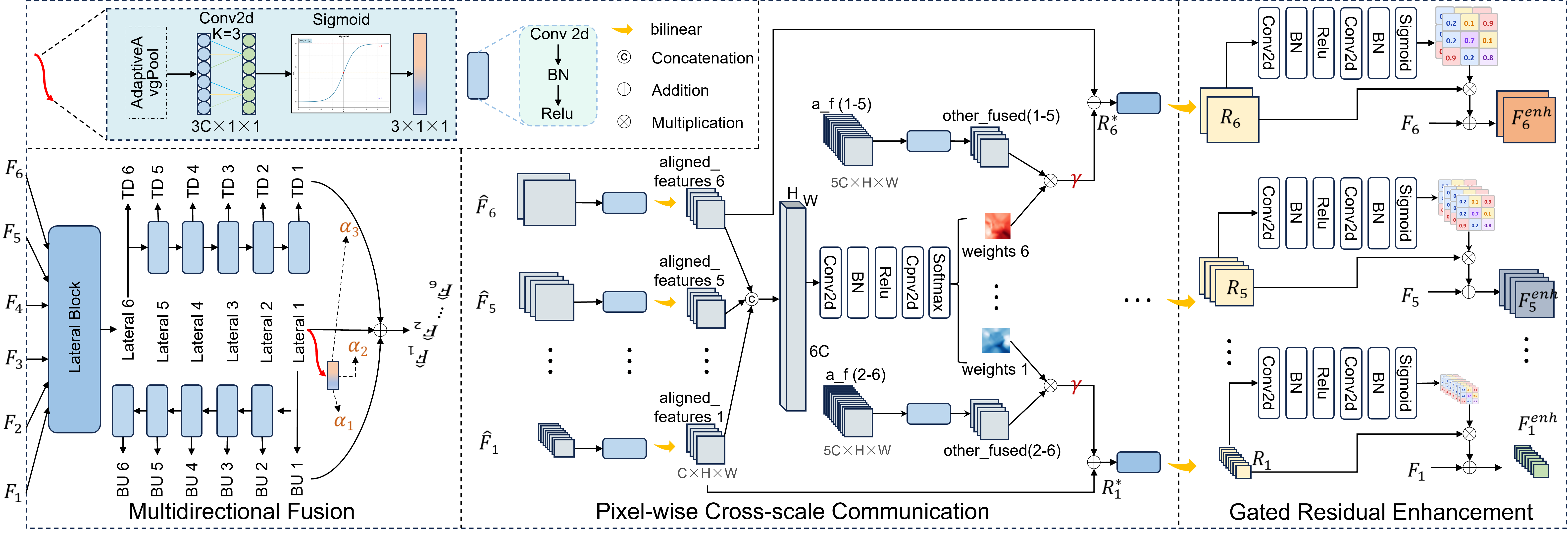}}
\caption{\textbf{Feature Communication Module (FCM).}
\emph{Left:} Multi-directional fusion—top–down (TD), bottom–up (BU), and lateral branches are built at every level; an SE head conditioned on $\mathbf{F}_i^{\text{lat}}$ produces weights $\boldsymbol{\alpha}_i$ that mix the three streams into $\widehat{\mathbf{F}}_i$.
\emph{Middle:} Pixel-wise cross-scale communication—all $\widehat{\mathbf{F}}_i$ are first projected with $1{\times}1$ convolutions and bilinearly aligned to a reference resolution $(H^\star{\times}W^\star)$, concatenated, and fed to the attention head $g$ to obtain per-pixel scale softmax weights $\mathbf{W}_{\text{att}}$. For each level $i$, features from the other scales are fused by a convolutional head and gated by $\mathbf{W}_{\text{att}}[i]$ to yield a per-level enhanced map $\mathbf{R}_i^\star$ at the reference resolution.
\emph{Right:} Gated residual enhancement—each $\mathbf{R}_i^\star$ is adapted by the mapping head $\varphi_i$ to $\mathbf{R}_i$, modulated by the sigmoid gate map $\mathbf{G}_i$, and fused residually with the original features $\mathbf{F}_i$ to produce $\mathbf{F}_i^{\text{out}}$.}

  \label{fig:fig2}
\end{figure*}
\section{Methodology}
\label{sec:Methodology}
We cast precipitation nowcasting as conditional spatiotemporal generation, where the model must capture multiscale echo evolution while remaining computationally efficient. We propose MFC-RFNet, an architecture that combines three complementary modules—FCM, WGSC, and CGSTF—and deploys VRWKV blocks at deep stages (the encoder tail, the bottleneck, and the first decoder layer) for efficient long-range temporal modeling. This section formalizes the task, outlines the overall network, and details each component.

\subsection{Problem setup and overall architecture}
Let $\mathbf{X}_t \in \mathbb{R}^{H\times W}$ denote a radar reflectivity (or rain-rate) image at time $t$. Given a history window $\mathbf{X}_{t-J+1:t}=\{\mathbf{X}_{t-J+1},\ldots,\mathbf{X}_t\}$, the goal is to predict the next $K$ frames $\mathbf{X}_{t+1:t+K}$. We model the conditional distribution
\begin{equation}
p\!\left(\mathbf{X}_{t+1:t+K}\mid \mathbf{X}_{t-J+1:t}\right),
\end{equation}
where the $K$ future frames are generated in a single shot as a spatiotemporal tensor $\mathbf{Z}\in\mathbb{R}^{K\times H\times W}$ (equivalently, $K$ channels).

\paragraph{Network Architecture}
The generator adopts a U-KAN \cite{li2025u} backbone, which is based on a four-scale U-Net architecture. A conditional encoder processes $\mathbf{X}_{t-J+1:t}$ and produces multi-scale features $\{\mathbf{F}_i\}_{i=1}^L$ (from shallow to deep). FCM performs bidirectional cross–scale communication before decoding. CGSTF operates at shallow stages (stages 1–3) to reduce inter-frame misalignment by predicting displacement fields from the conditioning branch and aligning backbone features prior to fusion. WGSC acts on deep and bottleneck skip connections, using wavelet-derived gates from the conditional stream to modulate encoder–decoder fusion. A compact VRWKV module is placed at the deepest stages—encoder tail, bottleneck, and first decoder layer—to supply long-range spatiotemporal context with moderate compute. See Fig.~\ref{fig:fig1} for the full schematic.

\subsection{Feature Communication Module (FCM)}
\textbf{Background and design rationale:}
Radar echo sequences contain broad motion patterns and rapidly evolving local echoes. Shallow features emphasize spatial localization and detail, whereas deep features summarize global context and longer-range dependencies. Simple concatenation or summation across scales can leave these cues weakly coupled, making it difficult to propagate global context to fine-grained localization. To address this, we construct a full-scale communication hub: multi-directional pathways connect pyramid levels in top-down, bottom-up, and lateral directions, and pixel-wise cross-scale selection enables each location to draw complementary information from the most relevant scale. This design strengthens the conditional representation without altering the backbone depth or resolution. An overview of FCM is shown in Fig.~\ref{fig:fig2}.

To allow conditional information to guide generation effectively, the encoder should coordinate information across scales rather than rely on unidirectional flows. The FCM enriches each scale with context from the others via multi-directional pathways and pixel-wise cross-scale selection.

\paragraph{Multidirectional fusion}
Let $\mathbf{F}_i$ denote the feature map at scale $i$, where $i\in\{1,\ldots,L\}$ indexes pyramid levels from the shallowest ($i=1$) to the deepest ($i=L$). FCM constructs
\begin{align}
\mathbf{F}^{\text{td}}_i &= \mathcal{U}\!\big(\phi^{\text{td}}_i(\mathbf{F}^{\text{td}}_{i+1})\big),\quad
\mathbf{F}^{\text{td}}_L=\phi^{\text{lat}}_L(\mathbf{F}_L),\\
\mathbf{F}^{\text{bu}}_i &= \mathcal{D}\!\big(\phi^{\text{bu}}_{i-1}(\mathbf{F}^{\text{bu}}_{i-1})\big),\quad
\mathbf{F}^{\text{bu}}_1=\phi^{\text{lat}}_1(\mathbf{F}_1),\\
\mathbf{F}^{\text{lat}}_i &= \phi^{\text{lat}}_i(\mathbf{F}_i),
\end{align}
where $\phi^{(\cdot)}$ are Conv–BN–ReLU blocks with $1{\times}1$ or $3{\times}3$ kernels, with $\mathcal{U}$ and $\mathcal{D}$ denoting bilinear upsampling and downsampling, respectively. Three paths are combined per scale via learned weights $\boldsymbol{\alpha}_i\in\mathbb{R}^3$ from squeeze-and-excitation (SE) over $\mathbf{F}^{\text{lat}}_i$:
\begin{equation}
\widehat{\mathbf{F}}_i=\sum_{j\in\{\text{td,bu,lat}\}}\alpha_{i,j}\,\mathbf{F}^{j}_i,\quad
\boldsymbol{\alpha}_i=\text{softmax}\!\big(\psi(\mathbf{F}^{\text{lat}}_i)\big).
\end{equation}

\paragraph{Pixel-wise cross-scale communication}
All multi-directional features $\widehat{\mathbf{F}}_i$ are first aligned to a common reference resolution (the middle scale) via learned $1{\times}1$ projections and bilinear resizing, producing a list of aligned features $\{\widetilde{\mathbf{F}}_i\}_{i=1}^L$. These are concatenated and processed by an attention head $g$ to compute per-pixel scale weights:
\begin{equation}
\mathbf{W}_{\text{att}}=\text{softmax}_{\text{scale}}\!\Big(g\big(\text{Concat}(\widetilde{\mathbf{F}}_1,\ldots,\widetilde{\mathbf{F}}_L)\big)\Big)\in\mathbb{R}^{L\times H^\star\times W^\star}.
\end{equation}
Instead of computing a single consensus map, we facilitate direct, per-scale communication. For each reference feature map $\widetilde{\mathbf{F}}_i$, we aggregate all \emph{other} maps via a fusion convolution
\(
\mathbf{F}_{\text{other}, i}=\phi_{\text{other}}\!\big(\text{Concat}(\{\widetilde{\mathbf{F}}_j\}_{j\neq i})\big),
\)
and compute the enhanced feature residually:
\begin{equation}
\mathbf{R}^{\star}_i=\widetilde{\mathbf{F}}_i+\gamma\,\big(\mathbf{W}_{\text{att}}[i]\odot \mathbf{F}_{\text{other}, i}\big),
\end{equation}
where $\gamma$ is a learnable scalar. This yields a set $\{\mathbf{R}^{\star}_i\}_{i=1}^L$ of enhanced features, one per scale, all at the reference resolution.

\paragraph{Gated residual enhancement}
The list $\{\mathbf{R}^{\star}_i\}$ is routed back to each native level $i$ (with channels $C_i$ and spatial size $H_i{\times}W_i$). A projection head $\varphi_i$ adapts each $\mathbf{R}^{\star}_i$ to its native dimensionality:
\begin{equation}
\mathbf{R}_i=\varphi_i\!\big(\text{Resize}(\mathbf{R}^{\star}_i\!\to\! i)\big)\in\mathbb{R}^{C_i\times H_i\times W_i}.
\end{equation}
To avoid indiscriminate injection of cross-scale context, a pixel- and channel-wise gating branch (Conv$3{\times}3$–BN–ReLU followed by Conv$1{\times}1$–BN–Sigmoid) produces a dense mask $\mathbf{G}_i$ from $\mathbf{R}_i$:
\begin{equation}
\begin{split}
\mathbf{G}_i
&=\sigma\!\Big(\text{BN}\!\big(\text{Conv}_{1\times1}(\text{BN}(\text{ReLU}(\text{Conv}_{3\times3}(\mathbf{R}_i))))\big)\Big)\\
&\in[0,1]^{C_i\times H_i\times W_i},
\end{split}
\end{equation}
where $\sigma$ denotes the sigmoid. The final output features are obtained by gated residual fusion with the original (pre-FCM) maps:
\begin{equation}
\mathbf{F}^{\text{out}}_i=\mathbf{F}_i+\mathbf{G}_i\odot \mathbf{R}_i,
\end{equation}
where $\odot$ denotes element-wise multiplication. The gate $\mathbf{G}_i$ selectively admits beneficial cross-scale information at each location and channel, while the residual path preserves the discriminative content of $\mathbf{F}_i$.

\begin{figure}[htbp]
  \centering{\includegraphics[scale=0.38]{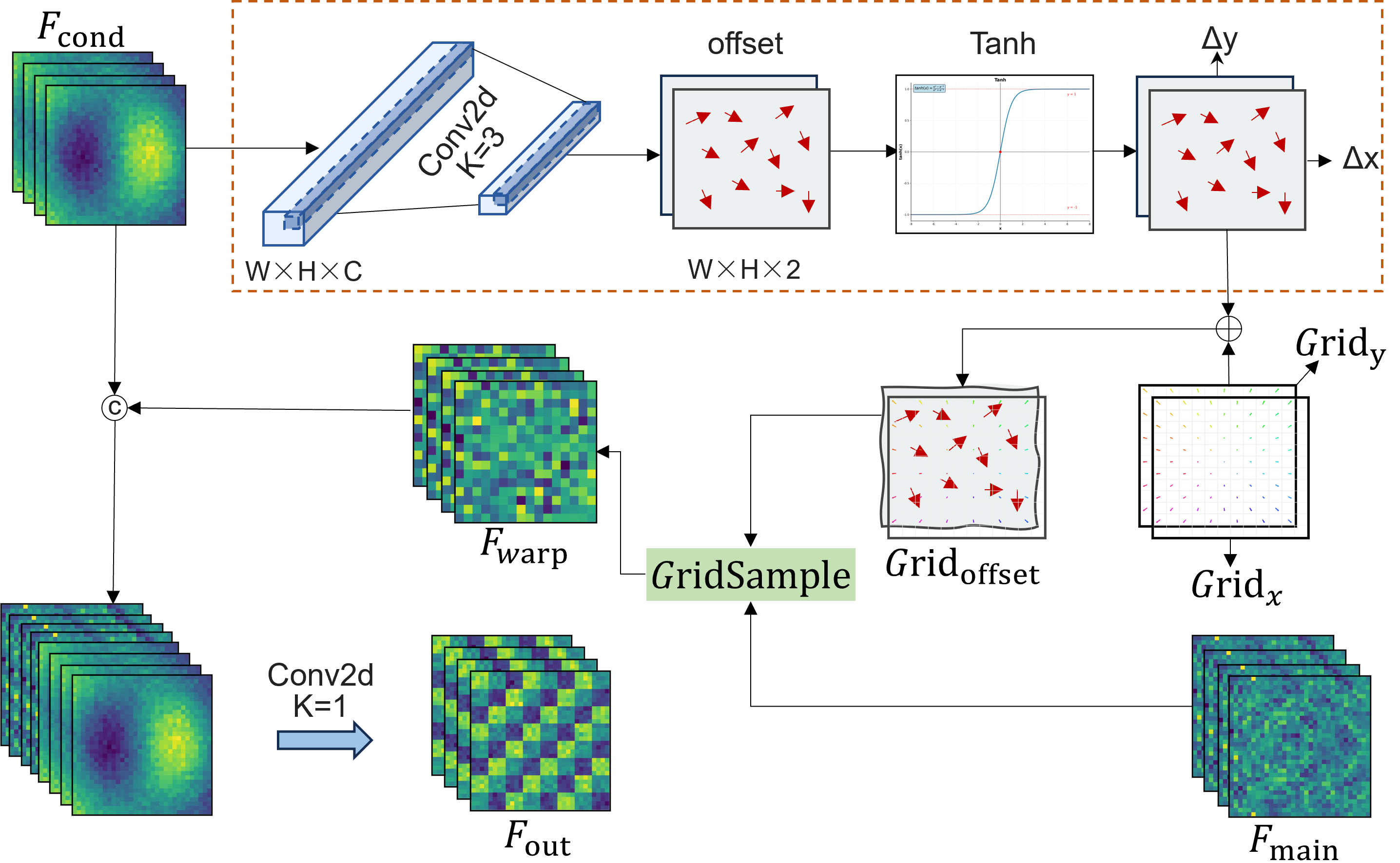}}
\caption{\textbf{Condition-Guided Spatial Transform Fusion (CGSTF).} Conditioning features produce an offset field $\mathbf{O}$ (bounded by $\tanh$) to perturb the base grid $(\mathrm{Grid}_x,\mathrm{Grid}_y)$ and obtain $\mathrm{Grid}_{\text{offset}}$. GridSample warps the main features to $\mathbf{F}_{\text{warp}}$, which is fused with the conditional features to yield $\mathbf{F}_{\text{out}}$.}
  \label{fig:fig3}
\end{figure}

\subsection{Condition-Guided Spatial Transform Fusion (CGSTF)}
\textbf{Background \& design rationale.}
The evolution of precipitation systems often involves complex, non-rigid motion, leading to significant spatial displacement between consecutive radar frames. Consequently, intermediate features within the network's main backbone can become spatially misaligned with those from the conditional encoder, particularly at shallow layers where features are still spatially explicit. Fusing these misaligned features directly can introduce spatial blurring and propagate localization errors to deeper, more abstract layers. A more robust strategy is to perform explicit alignment at these early stages. To this end, our CGSTF module learns to predict a displacement field directly from the conditioning echoes and applies a differentiable geometric transform, effectively warping the shallow backbone features to align them with the conditional features prior to fusion. To maintain training stability, the predicted offsets are bounded. An overview of CGSTF is shown in Fig.~\ref{fig:fig3}.\par

The alignment process begins by generating a displacement field $\mathbf{O}$ from the conditional features $\mathbf{F}_{\text{cond}}$ using a lightweight convolutional network $\phi$. The offsets are bounded for stability using a $\tanh$ function and a scalar hyperparameter $\alpha$:
\begin{equation}
\mathbf{O}=\alpha\,\tanh(\phi(\mathbf{F}_{\text{cond}}))\in\mathbb{R}^{B\times 2\times H\times W}.
\end{equation}
This field is permuted from \texttt{NCHW} to \texttt{NHWC} format ($\mathbf{O}_{\text{grid}}$) and added to a normalized base grid $\mathbf{G}_{\text{base}}\in[-1,1]^{B \times H\times W\times 2}$ to create the final sampling grid, $\mathbf{G}_{\text{offset}} = \mathbf{G}_{\text{base}} + \mathbf{O}_{\text{grid}}$. This grid is then used to warp the main backbone features $\mathbf{F}_{\text{main}}$ via differentiable bilinear sampling:
\begin{equation}
\mathbf{F}_{\text{warp}}
=\operatorname{GridSample}\!\big(\mathbf{F}_{\text{main}},\,\mathbf{G}_{\text{offset}}\big).
\end{equation}
The resulting warped features $\mathbf{F}_{\text{warp}}$ are now spatially aligned with the conditional features. They are concatenated and subsequently fused using a $1{\times}1$ convolution to mix information across channels:
\begin{equation}
\mathbf{F}_{\text{out}}
=\operatorname{Conv}_{1\times1}\!\big(\operatorname{Concat}(\mathbf{F}_{\text{warp}},\,\mathbf{F}_{\text{cond}})\big).
\end{equation}
This aligned feature map $\mathbf{F}_{\text{out}}$ is then passed to deeper stages. For implementation, coordinates in $\mathbf{G}_{\text{base}}$ are normalized to $[-1,1]$, and border padding is used in the $\operatorname{GridSample}$ operation to avoid edge artifacts. The bound $\alpha$ is set per stage, such that a unit offset approximately matches one pixel at that resolution. Deploying CGSTF at the shallowest stages (1–3) establishes geometric consistency early in the network, reducing error propagation and drift before high-level aggregation.

\begin{figure}[htbp]
  \centering{\includegraphics[scale=0.26]{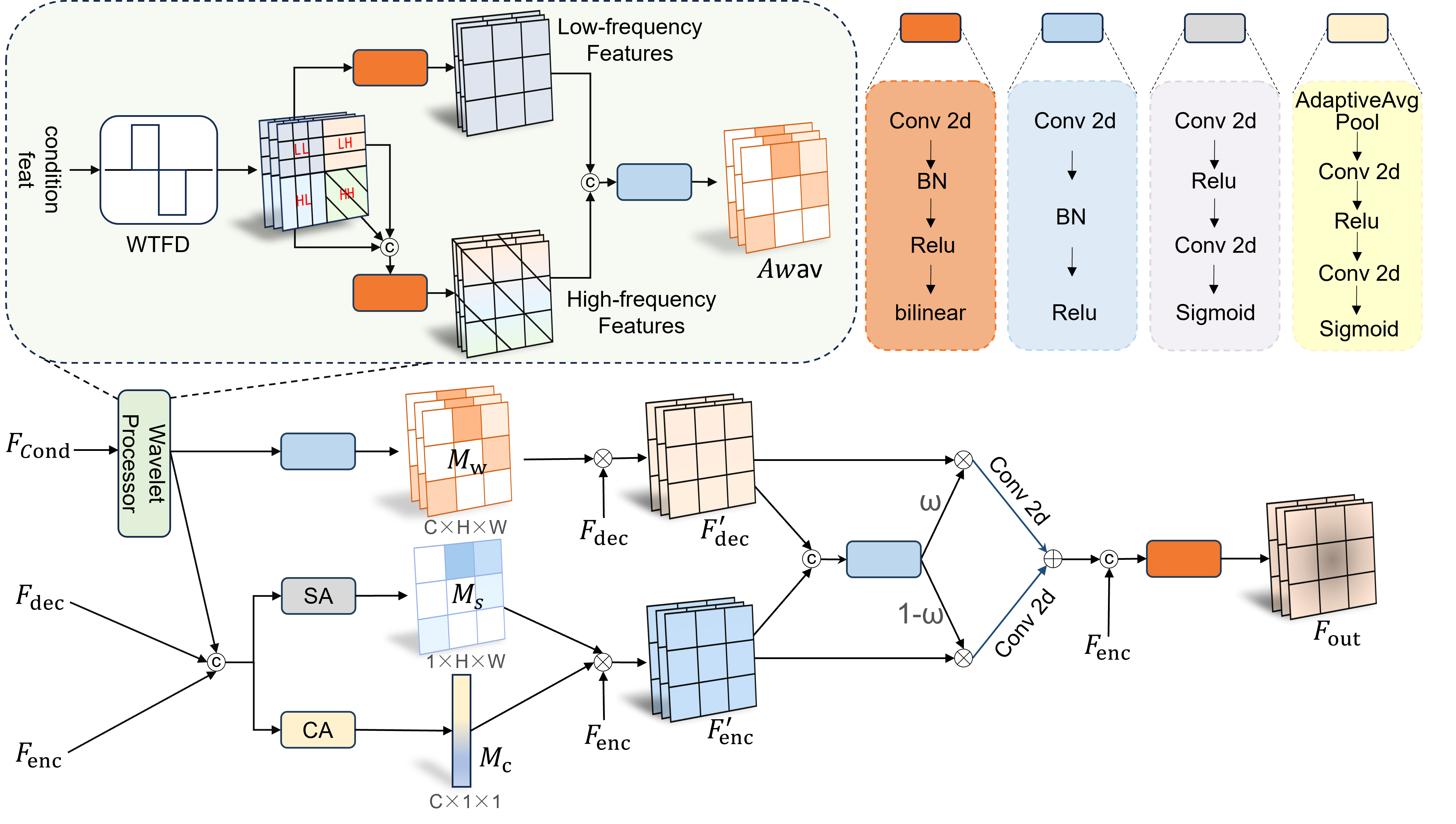}}
\caption{\textbf{The Wavelet-Guided Skip Connection (WGSC) architecture.} Conditional features are processed by the Wavelet Processor (top inset) via 2D DWT to yield a frequency-aware guidance map ($\mathbf{A}_{\text{wav}}$). Concurrently (main path, bottom), spatial (SA) and channel (CA) attention masks are derived from concatenated encoder and decoder features. These components, along with $\mathbf{A}_{\text{wav}}$, inform the synthesis of three distinct gates. Encoder features are modulated by SA/CA gates, while decoder features are modulated by a wavelet-derived gate. An adaptive fusion mechanism combines these modulated streams based on a learned weight $\omega$, followed by a final convolutional refinement step to produce the output.}
  \label{fig:fig4}
\end{figure}
\subsection{Wavelet-Guided Skip Connection (WGSC)}
In U-Net architectures, skip connections are crucial for propagating high-resolution details from the encoder to the decoder to aid reconstruction. However, standard skip connections are typically non-adaptive, passing information indiscriminately without regard to the current decoding context. This can introduce noise or irrelevant low-level features, potentially confounding the reconstruction. To make this fusion process more adaptive and context-aware, the WGSC leverages conditional features as external prior knowledge. Specifically, WGSC employs a 2D discrete wavelet transform (DWT) on the conditional stream to decompose features into different frequency sub-bands. This frequency-domain analysis provides a control mechanism to balance structural information (low-frequency) with fine-grained details (high-frequency) during the skip fusion process. An overview of WGSC's components is detailed below.

\paragraph{Wavelet guidance on the conditional branch}
Given conditional features $\mathbf{F}_{\text{cond}}$ at a deep/bottleneck scale, we first compute the 2D DWT (using \texttt{pywt} with 'db4'):
\begin{equation}
(\mathbf{F}_{LL},\mathbf{F}_{LH},\mathbf{F}_{HL},\mathbf{F}_{HH})=\text{DWT}_{2\text{D}}(\mathbf{F}_{\text{cond}}),
\end{equation}
where $LL$ encodes structural (low-frequency) information and $\{LH,HL,HH\}$ encode directional details. Two lightweight convolutional stems process the low-frequency and high-frequency components separately: $\mathbf{Q}_{\text{low}}=\phi_{\text{low}}(\mathbf{F}_{LL})$ and $\mathbf{Q}_{\text{high}}=\phi_{\text{high}}(\text{Concat}( \mathbf{F}_{LH},\mathbf{F}_{HL},\mathbf{F}_{HH}))$. Both $\mathbf{Q}_{\text{low}}$ and $\mathbf{Q}_{\text{high}}$ are bilinearly upsampled to the original resolution of $\mathbf{F}_{\text{cond}}$. These are then concatenated and passed through a final fusion block with a sigmoid activation to produce a unified wavelet guidance map, $\mathbf{A}_{\text{wav}}$:
\begin{equation}
\mathbf{A}_{\text{wav}}=\sigma\!\Big(\phi_{\text{fuse}}\big(\text{Concat}(\mathbf{Q}_{\text{low}}^{\uparrow},\mathbf{Q}_{\text{high}}^{\uparrow})\big)\Big)\in[0,1]^{C\times H\times W}.
\end{equation}
This map, $\mathbf{A}_{\text{wav}}$, encapsulates the frequency-domain information from the conditional features and serves as a key input for the subsequent gating stage.

\paragraph{Gate synthesis from conditional guidance}
The module then synthesizes three distinct gates by combining information from the encoder features ($\mathbf{F}_{\text{enc}}$), decoder features ($\mathbf{F}_{\text{dec}}$), and the newly computed wavelet guidance map ($\mathbf{A}_{\text{wav}}$). First, all three tensors are concatenated:
\begin{equation}
\mathbf{F}_{\text{comb}} = \text{Concat}(\mathbf{F}_{\text{enc}}, \mathbf{F}_{\text{dec}}, \mathbf{A}_{\text{wav}}).
\end{equation}
This combined tensor $\mathbf{F}_{\text{comb}}$ is processed by two parallel branches to derive context-aware masks: (i) a spatial attention (SA) module yielding a spatial mask $\mathbf{M}_s \in \mathbb{R}^{1 \times H \times W}$, and (ii) a channel attention (CA) module yielding a channel mask $\mathbf{M}_c \in \mathbb{R}^{C \times 1 \times 1}$. Concurrently, the wavelet guidance map $\mathbf{A}_{\text{wav}}$ is processed by its own lightweight network to produce a dedicated wavelet-based gate, $\mathbf{M}_w \in \mathbb{R}^{C \times H \times W}$. These three gates, $\{\mathbf{M}_s, \mathbf{M}_c, \mathbf{M}_w\}$, are then passed to the fusion stage.

\paragraph{Adaptive skip fusion}
The fusion process uses these three gates to selectively modulate the encoder and decoder features. The encoder feature $\mathbf{F}_{\text{enc}}$ is modulated by both the spatial and channel attention gates, while the decoder feature $\mathbf{F}_{\text{dec}}$ is modulated by the dedicated wavelet gate:
\begin{align}
\mathbf{F}'_{\text{enc}} &= \mathbf{F}_{\text{enc}} \odot \text{Broadcast}(\mathbf{M}_s) \odot \text{Broadcast}(\mathbf{M}_c) \\
\mathbf{F}'_{\text{dec}} &= \mathbf{F}_{\text{dec}} \odot \mathbf{M}_w
\end{align}
These two modulated features, $\mathbf{F}'_{\text{enc}}$ and $\mathbf{F}'_{\text{dec}}$, are then processed by separate $3 \times 3$ convolutions ($\phi_{\text{enc\_proc}}$ and $\phi_{\text{dec\_proc}}$) and fed into an adaptive fusion block. This block computes a dynamic fusion weight $\omega$ from their concatenation and performs a weighted sum:
\begin{equation}
\mathbf{F}_{\text{fused}} = \omega \odot \phi_{\text{enc\_proc}}(\mathbf{F}'_{\text{enc}}) + (\mathbf{1} - \omega) \odot \phi_{\text{dec\_proc}}(\mathbf{F}'_{\text{dec}}),
\end{equation}
where $\omega = \sigma(\phi_{\text{fusion}}(\text{Concat}(\phi_{\text{enc\_proc}}(\mathbf{F}'_{\text{enc}}), \phi_{\text{dec\_proc}}(\mathbf{F}'_{\text{dec}}))))$.
Finally, this result $\mathbf{F}_{\text{fused}}$ is concatenated with the original, unmodulated encoder feature $\mathbf{F}_{\text{enc}}$ and passed through a final convolutional block $\phi_{\text{out}}$ to produce the output $\mathbf{F}_{\text{out}}$:
\begin{equation}
\mathbf{F}_{\text{out}} = \phi_{\text{out}}(\text{Concat}(\mathbf{F}_{\text{fused}}, \mathbf{F}_{\text{enc}})).
\end{equation}
This multi-step, guided fusion allows the network to intelligently combine structural details from the encoder with context-aware decoder features, all under the guidance of frequency-domain information from the conditional input.

\textit{Implementation note:} in our implementation, the DWT is treated as a fixed analysis operator (no gradient through the transform), which stabilizes training while still learning the gates end-to-end. Placing DWT on the conditional/skip path lets the model modulate low-frequency structure and high-frequency detail during reconstruction, improving global consistency and edge sharpness while suppressing skip noise.

\begin{figure}[htbp]
  \centering{\includegraphics[scale=0.4]{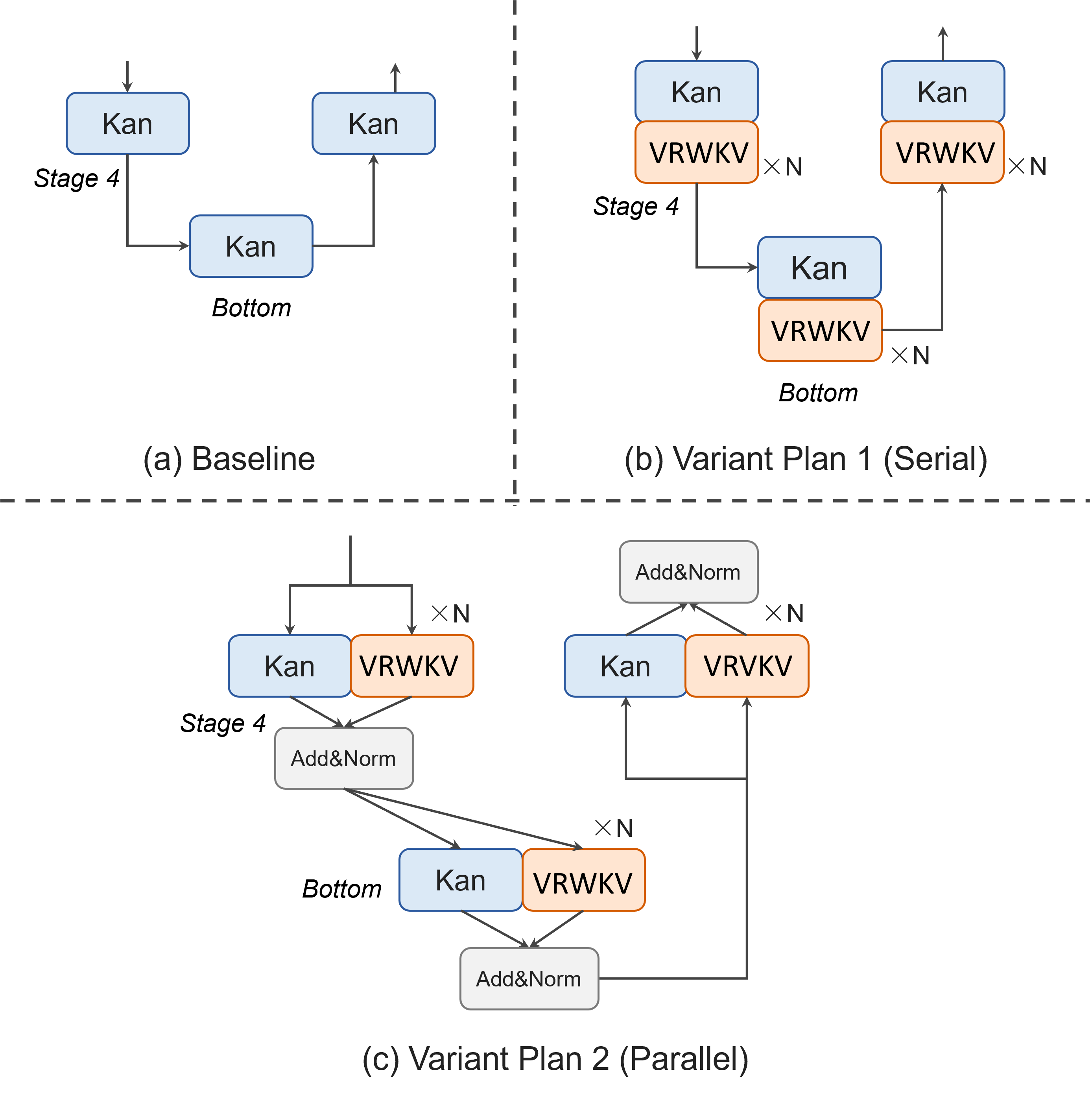}}
\caption{\textbf{VRWKV placement strategies at deep stages.} Comparison of integration strategies for VRWKV blocks at the encoder tail, bottleneck, and first decoder layer. \textbf{(a)} Baseline architecture at these locations, without VRWKV. \textbf{(b)} Serial insertion with $N$ blocks per stage; specifically, the $N=1$ case corresponds to the KAN-VRWKV Block configuration used in MFC-RFNet. \textbf{(c)} Parallel integration via residual mixing (Add\&Norm).}

  \label{fig:fig5}
\end{figure}
\subsection{VRWKV at deep stages for efficient long-range modeling}
\textbf{Background \& design rationale.}
As horizons and spatial resolution increase, quadratic self-attention over spatiotemporal tokens becomes a bottleneck. We place lightweight VRWKV blocks at deep stages—the encoder tail, bottleneck, and first decoder layer—where spatial resolution is lower and receptive fields are larger (with the bottleneck having the lowest resolution and the largest receptive field). VRWKV mixes tokens with linear-time sequence kernels while remaining parallel-trainable, extending temporal context at near-linear cost. We compared serial and parallel placements (Fig.~\ref{fig:fig5}); ablations favor a serial three-stage configuration across these deep locations as the best accuracy–efficiency trade-off (see Sec.~\ref{sec:experiments}).

\section{Experiments}
\label{sec:experiments}

\subsection{Datasets}
\label{sec:datasets}
Our proposed model is evaluated on four public radar precipitation datasets, chosen to cover diverse geographic regions and climatic conditions. For all datasets, we adopt a standard chronological split for training, validation, and testing sets to prevent temporal data leakage. All radar frames are resized to a uniform resolution of $128 \times 128$ pixels for processing. Unless otherwise specified, our models are trained to predict the next $K=20$ frames from the previous $J=5$ frames.

\paragraph{SEVIR}
The Storm Event Imagery Dataset (SEVIR) \cite{veillette2020sevir} contains 20,393 storm events over the Continental United States (CONUS) from 2017--2020, with a 5-minute cadence and a $384 \times 384$\,km spatial footprint. We use the vertically integrated liquid (VIL) product, where pixel values range from 0 to 255. Evaluation thresholds: $\{16, 74, 133, 160, 181, 219\}$.

\paragraph{MeteoNet}
This dataset \cite{larvor2020meteonet} provides radar data for northwestern France from 2016--2018, covering a $550 \times 550$\,km area at a 6-minute cadence. The reflectivity values range from 0 to 70\,dBZ, and the evaluation thresholds are $\{12, 18,
\\ 24, 32\}$\,dBZ.

\paragraph{Shanghai Radar}
This dataset \cite{chen2020deep} consists of radar echoes collected over Shanghai, China, from October 2015 to July 2018. It covers a $501 \times 501$\,km region with an approximate 6-minute cadence. The reflectivity range is 0--70\,dBZ, with evaluation thresholds of $\{20, 30, 35, 40\}$\,dBZ.

\paragraph{CIKM 2017 AnalytiCup}
This dataset \cite{yao2017cikm} covers a $101 \times 101$\,km area over Guangdong, China, with radar frames provided at 6-minute intervals. Reflectivity values range from 0 to 76\,dBZ, and thresholds are set at $\{20, 30, 35, 40\}$\,dBZ. For this specific dataset, we adjust the task to a 5$\rightarrow$10 frame prediction (a 60-minute forecast horizon).

\begin{table*}[htbp]
\centering
\caption{Experiment results on four radar datasets. The best results are highlighted in \textbf{bold} and the second-best results are \underline{underlined}.}
\label{tab:radar_results}
\resizebox{\textwidth}{!}{%
\begin{tabular}{l|cc|ccccc|ccccc}
\hline
& & & \multicolumn{5}{c|}{\textbf{SEVIR}} & \multicolumn{5}{c}{\textbf{MeteoNet}} \\
& Params(M)$\downarrow$ & GFLOPs$\downarrow$ & CSI-M$\uparrow$ & CSI-181$\uparrow$ & CSI-219$\uparrow$ & HSS$\uparrow$& MSE$\downarrow$ & CSI-M$\uparrow$ & CSI-24$\uparrow$ & CSI-32$\uparrow$ & HSS$\uparrow$& MSE$\downarrow$ \\
\hline
pySTEPS & -- & -- & 0.2830 & 0.1266 & \underline{0.0708} & 0.3673 & 652.83 & 0.3647 & 0.3552 & \underline{0.2273} & 0.4964 &16.42 \\
ConvGRU & 5.990 & 1962.556 & 0.2951 & 0.0846 & 0.0412 & 0.3696 & 369.72 & 0.3463 & 0.2911 & 0.1495 & 0.4741 & 13.67 \\
MAU & 51.218 & 291.462 & 0.3029 & 0.1129 & 0.0487 & 0.3799 & \underline{354.01} & 0.3162 & 0.2896 & 0.0920 & 0.4389 & 11.88 \\
SimVP & 11.038 & 51.089 & 0.3172 & 0.1061 & 0.0579 & 0.3988 & 382.44 & 0.3439 & 0.3078 & 0.1072 & 0.4516 & 14.82 \\
FourCastNet & 57.014 & 58.372 & 0.2621 & 0.0788 & 0.0287 & 0.3426 & 411.63 & 0.2969 & 0.2601 & 0.1146 & 0.4282 & 14.21 \\
Earthformer & 60.246 & 849.376 & 0.2837 & 0.0915 & 0.0211 & 0.3608 & 361.52 & 0.3276 & 0.2810 & 0.1299 & 0.4579 & 13.06 \\
PhyDNet & 5.829 & 286.214 & 0.3098 & 0.0987 & 0.0331 & 0.3749 & 356.02 & 0.3316 & 0.3269 & 0.1309 & 0.4604 & 15.96 \\
EarthFarseer & 94.093 & 907.237 & 0.2949 & 0.1058 & 0.0381 & 0.3907 & 389.88 & 0.3488 & 0.3093 & 0.1429 & 0.4802 & 12.69 \\
NowcastNet & 33.876 & 17.634 & 0.2868 & 0.0712 & 0.0397 & 0.3461  & 411.20 & 0.3360 & 0.3279 & 0.1669 & 0.4834 & 14.07 \\
DiffCast & 52.079 & 30616.382 & 0.3050 & 0.1300 & 0.0582 & 0.3996 & 559.59 & 0.3512 & 0.3340 & 0.1808 & 0.4846 & 17.93 \\
AlphaPre & 89.011 & 1550.903 & \underline{0.3259} & \underline{0.1332} & 0.0545 & \underline{0.4110} & \textbf{345.18} & \underline{0.3824} & \underline{0.3633} & 0.2002 & \underline{0.5164} & \underline{12.74} \\
MFC-RFNet(Ours) & 27.153 & 28.633& \textbf{0.3552} & \textbf{0.1966} & \textbf{0.1079} & \textbf{0.4576}  & 435.54 & \textbf{0.4213} & \textbf{0.4087} & \textbf{0.2477} & \textbf{0.5563}  & \textbf{11.91} \\
\hline
\hline
& & & \multicolumn{5}{c|}{\textbf{Shanghai}} & \multicolumn{5}{c}{\textbf{CIKM}} \\
& Params(M)$\downarrow$ & GFLOPs$\downarrow$ & CSI-M$\uparrow$ & CSI-35$\uparrow$ & CSI-40$\uparrow$ & HSS$\uparrow$& MSE$\downarrow$ & CSI-M$\uparrow$ & CSI-35$\uparrow$ & CSI-40$\uparrow$ & HSS$\uparrow$& MSE$\downarrow$ \\
\hline
pySTEPS & -- & -- & 0.3719 & 0.3376 & 0.2514 & 0.4995 & 42.11 & 0.2856 & \underline{0.2141} & \underline{0.1550} & 0.3779 & 80.83 \\
ConvGRU & 5.990 & 1962.556 & 0.3687 & 0.3096 & 0.2133 & 0.4950 & 32.17 & 0.3166 & 0.1942 & 0.1322 & 0.3927 & 38.94 \\
MAU & 51.218 & 291.462 & 0.3904 & 0.3698 & 0.2350 & 0.5289 & 31.92 & 0.2976 & 0.2120 & 0.1185 & 0.3991 & 39.65 \\
SimVP & 11.038 & 51.089 & 0.3922 & 0.3471 & 0.2448 & 0.5267 & 35.86 & 0.3139 & 0.1972 & 0.1388 & 0.4046 & 36.75 \\
FourCastNet & 57.014 & 58.372 & 0.3506 & 0.3189 & 0.2001 & 0.4795 & \underline{31.05} & 0.2905 & 0.1926 & 0.0959 & 0.3874 & 37.56 \\
Earthformer & 60.246 & 849.376 & 0.3589 & 0.3102 & 0.1957 & 0.4928 & 33.94 & 0.3001 & 0.2118 & 0.1308 & 0.4078 & \underline{36.07} \\
PhyDNet & 5.829 & 286.214 & 0.3599 & 0.3319 & 0.2108 & 0.4880 & 37.62 & 0.3109 & 0.1989 & 0.1343 & 0.3864 & 40.93 \\
EarthFarseer & 94.093 & 907.237 & 0.4011 & 0.3534 & 0.2420 & 0.5409 & 34.23 & 0.2983 & 0.2112 & 0.1317 & 0.3984 & 38.22 \\
NowcastNet & 33.876 & 17.634 & 0.3886 & 0.3681 & 0.2389 & 0.5405  & 34.77 & 0.3074 & 0.1879 & 0.1269 & 0.3939  & 39.51 \\
DiffCast & 52.079 & 30616.382 & 0.4089 & 0.3740 & 0.2606 & 0.5476  & 36.35 & 0.3159 & 0.2009 & 0.1457 & 0.4085 & 42.78 \\
AlphaPre & 89.011 & 1550.903 & \underline{0.4178} & \underline{0.3854} & \underline{0.2615} & \underline{0.5534}  & \textbf{28.02} & \underline{0.3194} & 0.2068 & 0.1416 & \underline{0.4137}  & \textbf{35.18} \\
MFC-RFNet(Ours) & 27.153 & 28.633 & \textbf{0.4198} & \textbf{0.3867} & \textbf{0.2823} & \textbf{0.5583}  & 32.91 & \textbf{0.3292} & \textbf{0.2406} & \textbf{0.1666} & \textbf{0.4278}  & 43.16 \\
\hline
\end{tabular}%
}
\vspace{2mm}
\noindent\footnotesize Note: “Params (M)” and “GFLOPs” were computed assuming an input of $(1,5,1,128,128)$ and an output of $(1,20,1,128,128)$.
\end{table*}

\subsection{Implementation Details}
\label{sec:impl}

\paragraph{Training and Model Selection}
All models were trained for 500 epochs using the AdamW optimizer with an initial learning rate of $1\times10^{-4}$ and a cosine annealing schedule. We used a batch size of 8 and applied exponential moving average (EMA) with a decay of 0.95 to the model weights. For preprocessing, all input frames were resized to $128\times128$ and linearly normalized to the $[0,1]$ range based on each dataset’s native value scale. The optimization objective is the RF loss only. For reproducibility, the random seed was fixed to 0. The final model for testing was selected based on the best mean Critical Success Index (CSI-M) on the validation set. During inference, we used a fixed 5-step ODE sampler to generate forecasts.

\paragraph{Architecture Details}
Our MFC-RFNet architecture consists of a U-KAN generative backbone and a separate conditional encoder. Concretely, U-KAN \cite{li2025u} preserves the classic U-Net encoder–decoder with skip connections while inserting a tokenized KAN block near the bottleneck to project features into tokens and apply the KAN operator, thereby strengthening nonlinear modeling capacity and interpretability. The backbone features a four-level hierarchical pyramid to handle multi-scale information. The conditional encoder, which shares the same encoder structure as the backbone, processes the input sequence to produce multi-scale conditional features ${\mathbf{F}_i}$. These features are first enhanced by the FCM module through multi-directional pathways and pixel-wise attention. The resulting features ${\mathbf{F}^{\text{enh}}_i}$ are then injected into the corresponding levels of the main backbone's decoder. Within the backbone, the CGSTF module is applied at the three shallowest encoder stages (1–3) to perform explicit feature alignment and reduce motion-induced misalignment. In the decoder, the WGSC module operates on the two deepest skip connections, using wavelet-derived cues to intelligently modulate the fusion of encoder–decoder features. Finally, we insert VRWKV blocks at three deep stages of the backbone in a serial manner—one at the encoder tail, one at the bottleneck, and one at the first decoder layer. Unless otherwise noted, all experiments adopt this serial one-block, three-stage configuration. Each VRWKV uses a 1/4 channel compression ratio and a pixel shift of 1, consistent with our ablation settings.

\paragraph{Environment}
All experiments were implemented in PyTorch 2.2.2 on an Ubuntu 22.04.1 system with CUDA 12.1. The models were trained and evaluated on a single NVIDIA RTX 4090 GPU.

\begin{figure}[htbp]
    \centering

    \includegraphics[width=0.48\textwidth]{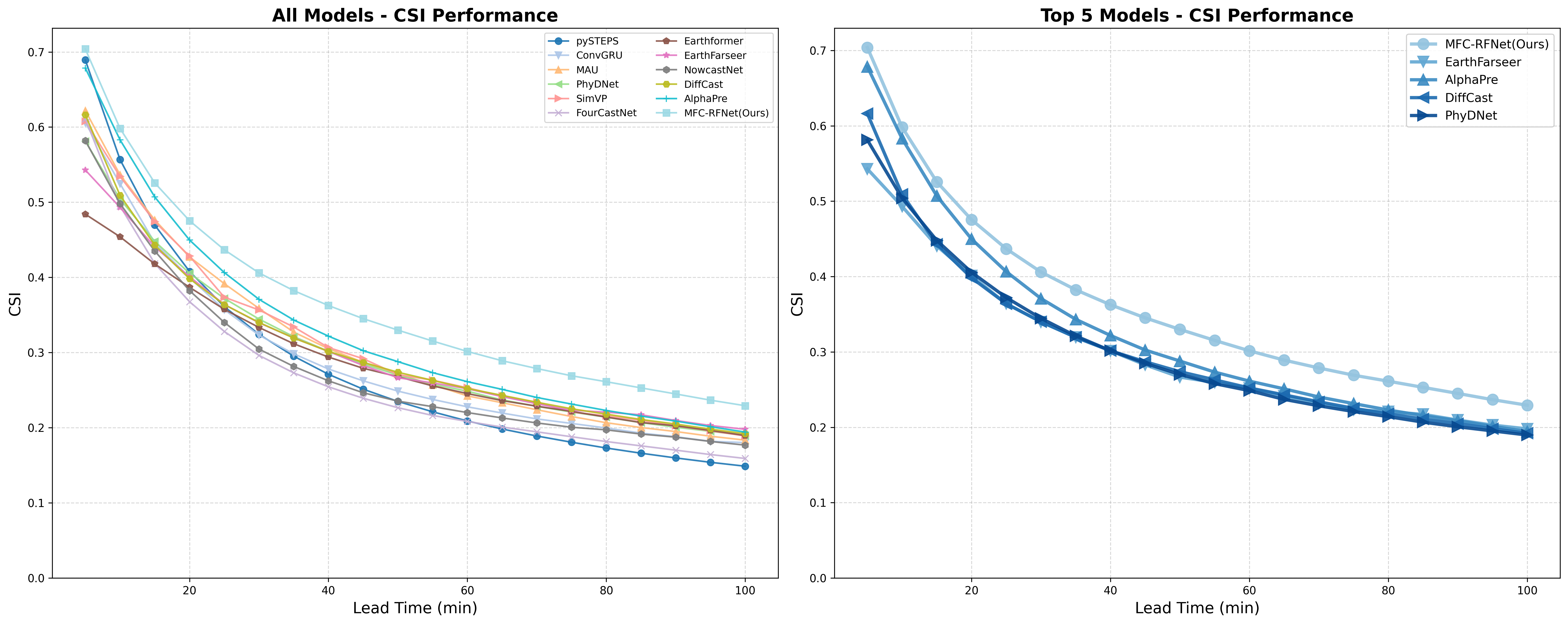}
    \label{fig:csi_metrics}

    \vspace{0.5cm}  

    \includegraphics[width=0.48\textwidth]{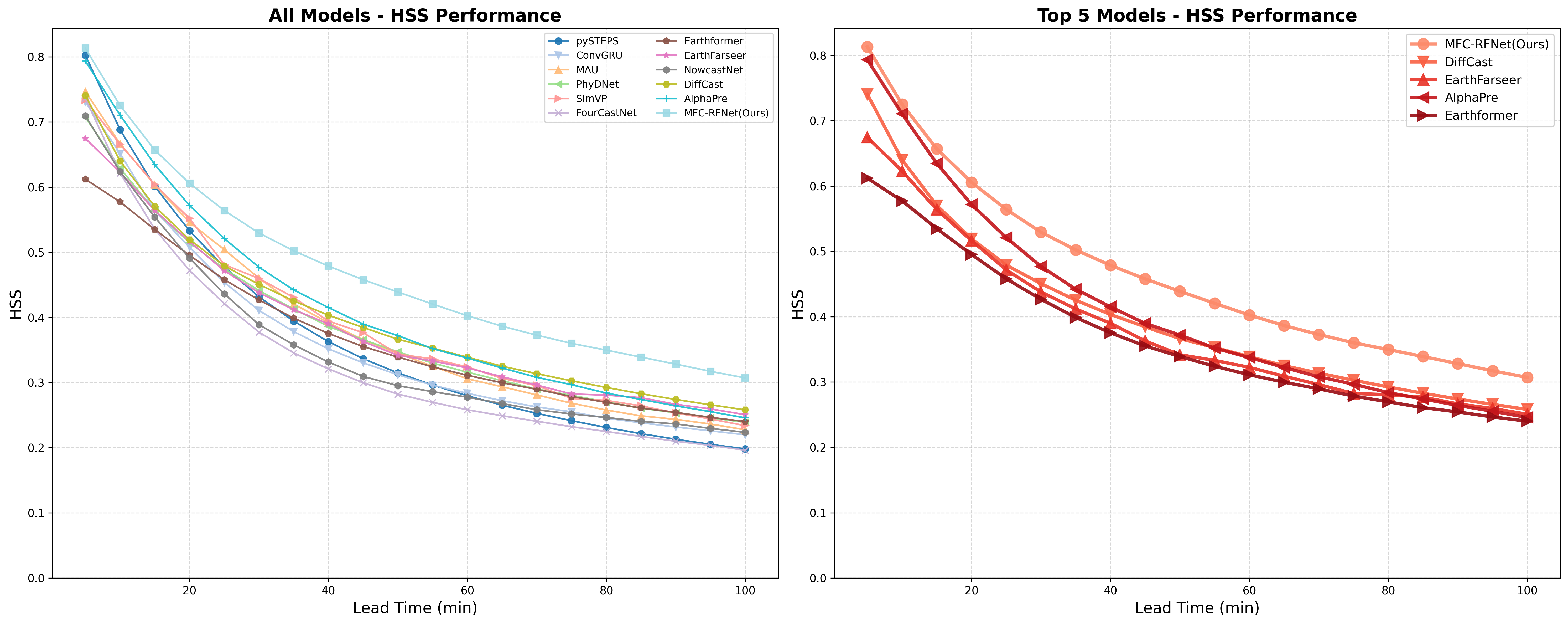}
    \label{fig:hss_metrics}

    \caption{CSI and HSS metrics at different prediction time steps for various methods on the SEVIR dataset.}
    \label{fig:csi_hss_metrics}
\end{figure}

\begin{table*}[htbp]
\centering
\caption{Ablation study results on four datasets. We progressively add each component to demonstrate their individual contributions. The best results are shown in \textbf{bold}.}
\label{tab:ablation}
\resizebox{\textwidth}{!}{
\begin{tabular}{l|ccccc|ccccc}
\hline
\toprule
\multirow{2}{*}{Method} & \multicolumn{5}{c|}{SEVIR} & \multicolumn{5}{c}{MeteoNet} \\
 & CSI-M↑ & CSI-181↑ & CSI-219↑ & HSS↑ & MSE↓ & CSI-M↑ & CSI-24↑ & CSI-32↑ & HSS↑ & MSE↓ \\
 \hline
\midrule
backbone(U-KAN) & 0.3095 & 0.1049 & 0.0401 &0.3875 & 461.26 &0.3793 & 0.3453 & 0.1601 & 0.5026 & 13.31 \\
+ RF & 0.3456 & 0.1823 & 0.0912 & 0.4234 & 451.23 & 0.4021 & 0.3834 & 0.2298 & 0.5387 & 12.67 \\
+ CGSTF & 0.3389 & 0.1887 & 0.0945 & 0.4312 & 448.91 & 0.4156 & 0.3956 & 0.2401 & 0.5456 & 12.34 \\
+ WGSC & 0.3478 & 0.1834 & 0.0978 & 0.4287 & 442.16 & 0.4087 & 0.3912 & 0.2356 & 0.5423 & 12.18 \\
+ FCM & 0.3521 & 0.1945 & 0.1023 & 0.4456 & 439.87 & 0.4198 & 0.4034 & 0.2445 & 0.5512 & 12.03 \\
+ VRWKV & \textbf{0.3552} & \textbf{0.1966} & \textbf{0.1079} & \textbf{0.4576} & \textbf{435.54} & \textbf{0.4213} & \textbf{0.4087} & \textbf{0.2477} & \textbf{0.5563} & \textbf{11.91} \\
\hline
\bottomrule
\end{tabular}
}
\vspace{0.3cm}
\resizebox{\textwidth}{!}{
\begin{tabular}{l|ccccc|ccccc}
\hline
\toprule
\multirow{2}{*}{Method} & \multicolumn{5}{c|}{Shanghai} & \multicolumn{5}{c}{CIKM} \\
 & CSI-M↑ & CSI-35↑ & CSI-40↑ & HSS↑ & MSE↓ & CSI-M↑ & CSI-35↑ & CSI-40↑ & HSS↑ & MSE↓ \\
\hline
\midrule
backbone(U-KAN) & 0.4002 & 0.3550 & 0.2420 & 0.5260 & 34.01 & 0.2988 & 0.2023 & 0.1276 & 0.3849 & 46.99 \\
+ RF & 0.4087 & 0.3734 & 0.2687 & 0.5345 & 34.78 & 0.3098 & 0.2134 & 0.1389 & 0.3912 & 46.23 \\
+ CGSTF & 0.4023 & 0.3698 & 0.2612 & 0.5289 & 34.92 & 0.3067 & 0.2098 & 0.1356 & 0.3876 & 45.87 \\
+ WGSC & 0.4156 & 0.3821 & 0.2756 & 0.5467 & 33.89 & 0.3156 & 0.2198 & 0.1434 & 0.4012 & 45.12 \\
+ FCM & 0.4134 & 0.3789 & 0.2734 & 0.5423 & 33.65 & 0.3234 & 0.2334 & 0.1567 & 0.4187 & 44.34 \\
+ VRWKV & \textbf{0.4198} & \textbf{0.3867} & \textbf{0.2823} & \textbf{0.5583} & \textbf{32.91} & \textbf{0.3293} & \textbf{0.2406} & \textbf{0.1666} & \textbf{0.4278} & \textbf{43.16} \\
\hline
\bottomrule
\end{tabular}
}
\end{table*}

\subsection{Quantitative Evaluation}
\label{sec:quantitative_evaluation}
We benchmark MFC\hbox{-}RFNet against a representative set of baselines spanning extrapolation and learning paradigms: pySTEPS \cite{pulkkinen2019pysteps}; recurrent models (ConvGRU \cite{wang2018learning}, MAU \cite{chang2021mau}); efficient video predictors (SimVP \cite{gao2022simvp}, FourCastNet \cite{pathak2022fourcastnet}); Transformer families (EarthFormer \cite{gao2022earthformer}, EarthFarseer \cite{wu2024earthfarsser}); a physics\hbox{-}informed design (PhyDNet \cite{guen2020disentangling}); a nowcasting\hbox{-}specific model (AlphaPre \cite{lin2025alphapre}); and recent generative approaches (NowcastNet \cite{zhang2023skilful}, DiffCast \cite{yu2024diffcast}). Metrics include CSI at multiple thresholds (reported as CSI\hbox{-}M and per\hbox{-}threshold CSI), HSS, and MSE.

Table~\ref{tab:radar_results} summarizes results on four datasets. MFC-RFNet attains the best CSI-M and HSS on each dataset while keeping MSE competitive, indicating improvements that hold across climate regimes and evaluation ranges. On SEVIR, CSI-M is 0.3552 with HSS 0.4576—about nine and eleven percent higher than the next best, respectively. MeteoNet and Shanghai show similar trends, and on CIKM the model maintains a consistent edge despite the shorter prediction horizon. Performance at higher rain-rate thresholds is particularly important: on SEVIR, CSI-219 reaches 0.1079 ($\approx$85\% higher than the strongest generative baseline, DiffCast), and on MeteoNet, CSI-32 also shows a clear margin. Notably, the traditional pySTEPS baseline remains highly competitive and even outperforms most deep-learning counterparts at the highest thresholds (e.g., SEVIR CSI-219, MeteoNet CSI-32, CIKM CSI-40), highlighting the persistent challenge of forecasting intense convection. Nevertheless, our model consistently and significantly surpasses pySTEPS in these critical regimes, indicating that the design remains effective when evaluation focuses on intense echoes, where many methods tend to underestimate and oversmooth.

Fig.~\ref{fig:csi_hss_metrics} reports lead-time curves on SEVIR. Across 5 to 100 forecast steps, the CSI and HSS curves of MFC-RFNet remain above all baselines. The gap is most visible after about 30 steps, where many methods decay rapidly while our curve exhibits a slower decline, suggesting that cross-scale communication and frequency-guided fusion help retain structure as the horizon grows. The right panels (top five models) highlight that the advantage is not restricted to a single regime: both early lead times and late lead times benefit, with the largest margins appearing at longer horizons.

\begin{figure*}[htbp]
  \centering{\includegraphics[scale=0.64]{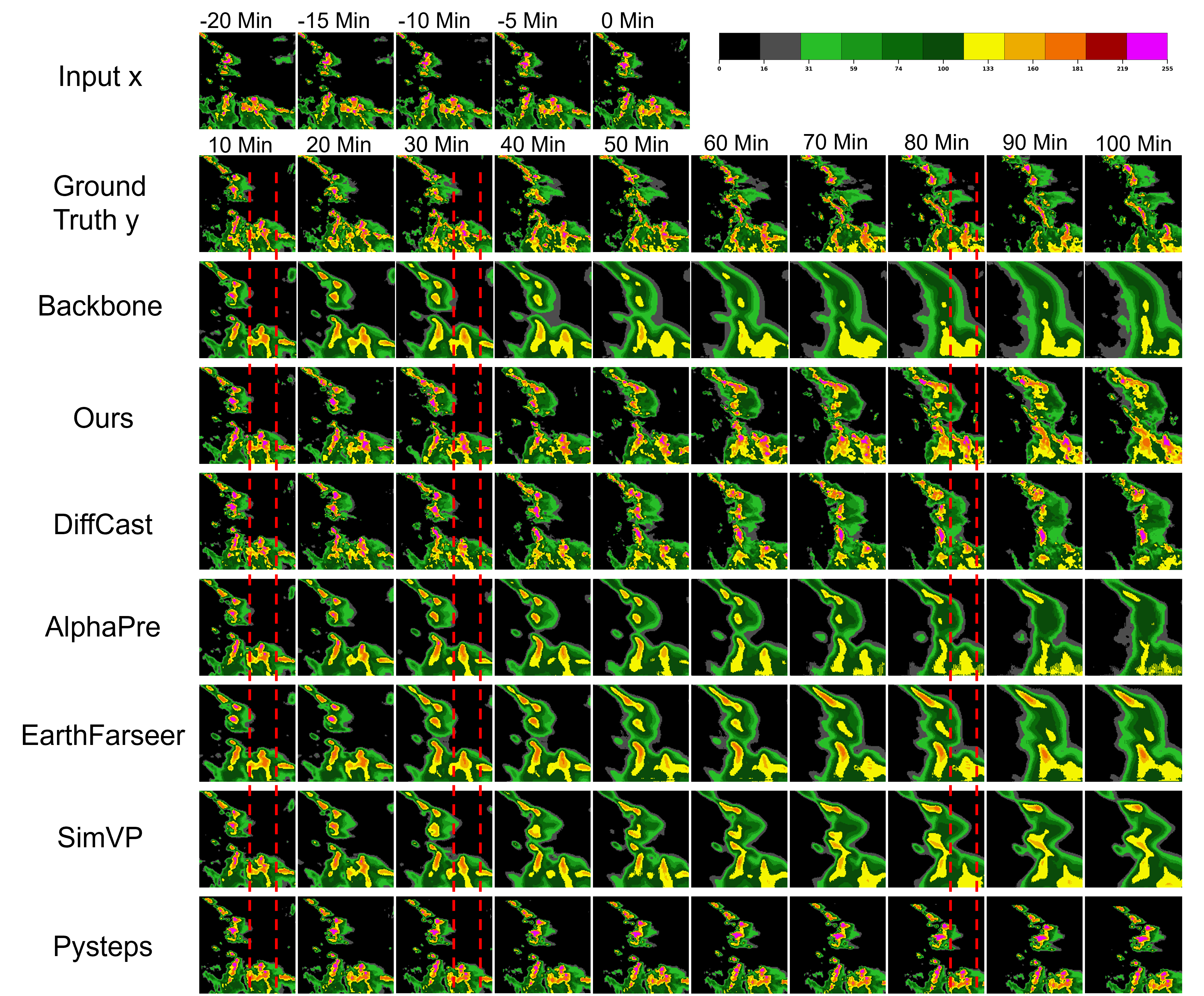}}
\caption{\textbf{Qualitative comparisons on the SEVIR dataset.}}
  \label{fig:qualitativesevir}
\end{figure*}

\begin{figure*}[htbp]
  \centering{\includegraphics[scale=0.64]{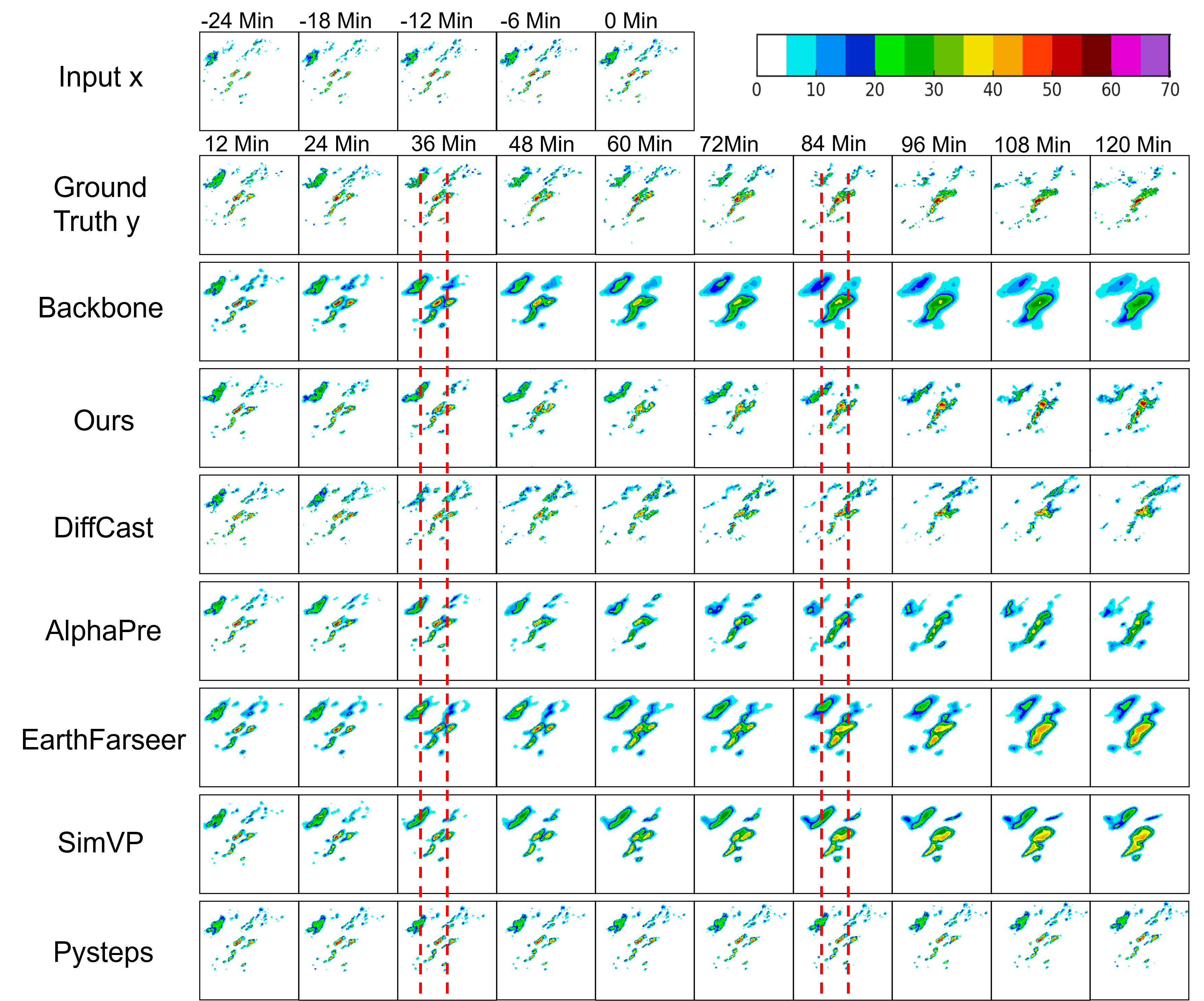}}
\caption{\textbf{Qualitative comparisons on the Shanghai dataset.}}
  \label{fig:qualitativeShanghai}
\end{figure*}

\begin{figure*}[htbp]
  \centering{\includegraphics[scale=0.64]{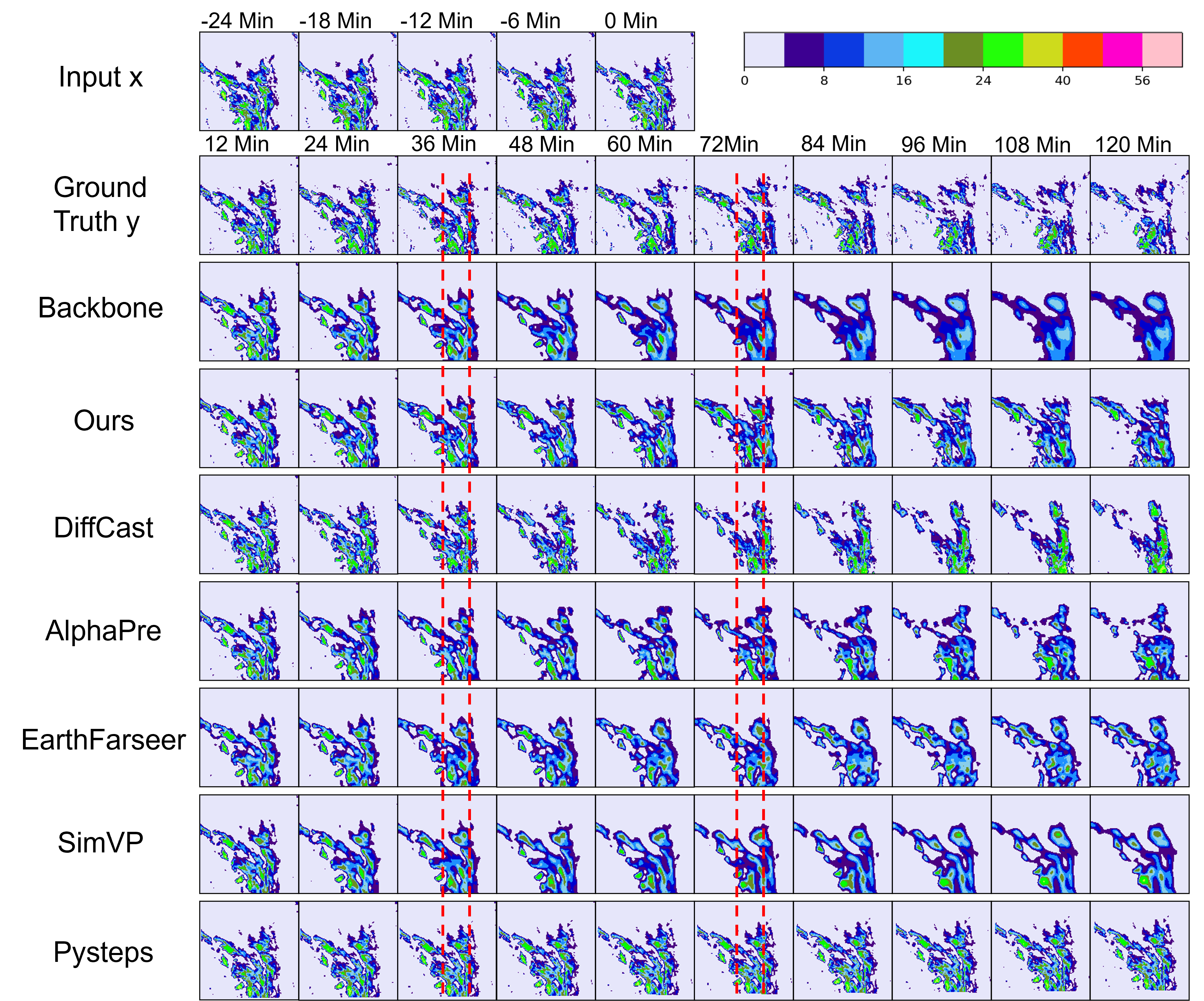}}
\caption{\textbf{Qualitative comparisons on the MeteoNet dataset.}}
  \label{fig:qualitativeMeteoNet}
\end{figure*}

\begin{figure*}[htbp]
  \centering{\includegraphics[scale=0.64]{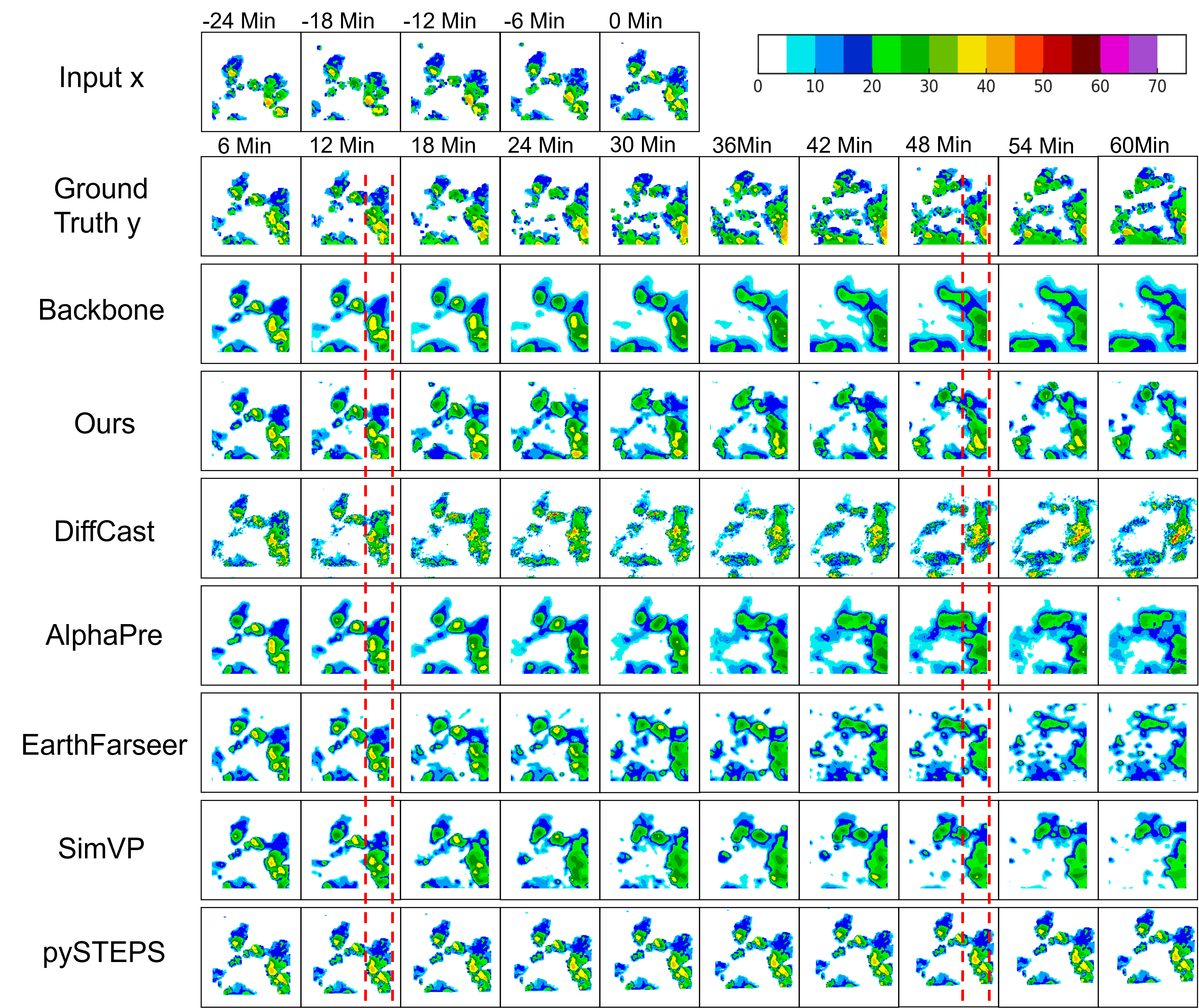}}
\caption{\textbf{Qualitative comparisons on the CIKM dataset.}}
  \label{fig:qualitativeCIKM}
\end{figure*}

Fig.~\ref{fig:qualitativesevir}--\ref{fig:qualitativeCIKM} present visual comparisons at multiple forecast lead times across the four datasets (SEVIR, Shanghai, MeteoNet, CIKM). Relative to the baselines, MFC-RFNet exhibits less attenuation of strong echoes and reduced spatial diffusion as the horizon increases, with clearer echo edges and more coherent morphology. The effect is most visible at later lead times (for example on SEVIR beyond 60 minutes), where reference peak locations and fine-scale gradients are better preserved and spurious weak echoes are reduced. On Shanghai and MeteoNet, narrow bands remain more continuous; on CIKM, small cells show fewer fragmented artifacts. These observations are consistent with the intended roles of FCM, CGSTF, and WGSC in coordinating scales, aligning shallow features, and modulating skip fusion.

From a cost perspective, the model uses about 27 million parameters and 28.6 GFLOPs in the reported configuration (see Table~\ref{tab:radar_results}), which is comparable to or smaller than several strong baselines while delivering higher skill. Taken together, the experiment results on four radar datasets, the CSI and HSS metrics evaluated at different prediction time steps on the SEVIR dataset, and qualitative visual comparisons across datasets provide consistent evidence that combining rectified flow training with scale-aware communication, shallow alignment, and wavelet-guided skip modulation yields more stable skill over time and more reliable detection at high thresholds, without incurring excessive computational cost.


\subsection{Ablation Studies}
\label{sec:ablation_studies}

To meticulously validate the contribution of each architectural innovation, we conducted a series of extensive ablation studies. These studies are designed to first demonstrate the cumulative benefit of our proposed components and then to dissect the internal mechanisms of each key module.

\subsubsection{Analysis of Progressive Module Integration}
\label{sec:ablation_progressive}

Table~\ref{tab:ablation} presents a progressive integration study, starting from a baseline backbone and incrementally adding each core component of MFC-RFNet. The initial transition to an RF training paradigm provides the most significant single performance boost across all datasets, particularly at higher intensity thresholds on SEVIR. This confirms that the near-linear probability trajectories learned by RF are highly effective for preserving the fidelity of strong echoes.

Subsequently, introducing CGSTF demonstrates dataset-specific effects and modular synergy. On MeteoNet, which features fast-moving, small-scale echoes, it provides a clear performance gain (Table~\ref{tab:ablation}). This result is consistent with its intended role: by performing explicit alignment at shallow layers, CGSTF effectively mitigates motion-induced misalignments before they can propagate and degrade deeper features. On SEVIR, a slight performance dip is observed when CGSTF is added at this stage, but this is immediately recovered and surpassed by the subsequent addition of WGSC. This pattern suggests a strong synergistic interaction, where the benefits of shallow alignment (CGSTF) are more fully realized when combined with the frequency-aware skip fusion (WGSC). The addition of WGSC raises CSI scores at medium and high thresholds while slightly lowering MSE, suggesting that its frequency-aware gating mechanism successfully preserves important structural details during reconstruction. Incorporating FCM improves cross-scale coherence, yielding steady gains on CIKM and Shanghai and slowing skill decay at longer lead times. Finally, deploying VRWKV blocks serially at the encoder tail, the bottleneck, and the first decoder layer provides a modest yet consistent further improvement, confirming the benefit of augmenting low-resolution deep stages with long-range context at minimal overhead.

\begin{table}[htbp]
\centering
\scriptsize
\caption{Ablation study on FCM components on MeteoNet dataset. We analyze the contribution of each component within the FCM module. The best results are shown in \textbf{bold}.}
\label{tab:fcm_components3}
\setlength{\tabcolsep}{5pt}
\begin{tabular}{l|ccccc}
\hline
\toprule
Method & CSI-M↑ & CSI-24↑ & CSI-32↑ & HSS↑ & MSE↓ \\
\hline
\midrule
w/o Cross Attention & 0.4089 & 0.3945 & 0.2334 & 0.5412 & 12.43 \\
w/o Multi Fusion & 0.4034 & 0.3889 & 0.2298 & 0.5367 & 12.56 \\
w/o CA \& MF & 0.3987 & 0.3823 & 0.2245 & 0.5298 & 12.78 \\
w/o Enhancement& 0.4178 & 0.4045 & 0.2423 & 0.5521 & 12.08 \\
\midrule
Full FCM & \textbf{0.4213} & \textbf{0.4087} & \textbf{0.2477} & \textbf{0.5563} & \textbf{11.91} \\
\hline
\bottomrule
\end{tabular}
\end{table}

\subsubsection{Dissection of the Feature Communication Module (FCM)}
\label{sec:ablation_fcm}

The FCM consists of three main components: multi-directional fusion, cross-scale attention, and gated residual enhancement. As shown in Table~\ref{tab:fcm_components3} on the MeteoNet dataset, removing any component degrades performance, confirming their complementary roles. Disabling multi-directional fusion causes the most significant drop in high-threshold CSI (a decrease of about 8\% for CSI-32), highlighting the critical importance of bidirectional information flow for coordinating global context and local details. Removing the cross-scale attention mechanism also leads to a notable performance decline, particularly for high-intensity events, which indicates that the per-pixel adaptive scale selection is crucial for retaining sharp, well-defined structures. The final enhancement stage provides a smaller but consistent benefit, helping to effectively integrate the globally-aware features back into each scale. The full FCM configuration achieves the best overall performance, supporting our hypothesis that a holistic communication strategy is superior to simpler fusion methods.

\subsubsection{Analysis of the Conditional-Guided Spatial Transform Fusion (CGSTF)}
\label{sec:ablation_cgstf}

We evaluated alternative fusion strategies to CGSTF on the SEVIR dataset, with results in Table~\ref{tab:cgstf_components4}. Replacing our explicit alignment-and-fusion approach with simple element-wise addition or concatenation leads to a significant drop in performance, especially at the highest threshold (CSI-219 decreases by 8.5\% and 6.2\%, respectively). This confirms that directly merging spatially misaligned features is suboptimal. The most substantial degradation occurs when the offset prediction network is removed entirely (CSI-219 drops by 10.4\%), underscoring that the learned, data-driven displacement field is the key mechanism for improving high-threshold skill. Furthermore, removing the tanh activation, which bounds the predicted offsets, also results in a consistent, albeit smaller, performance drop, suggesting that this constraint is important for maintaining stable and physically plausible warping. These findings collectively validate that the explicit, bounded, and condition-guided warping performed by CGSTF is a more effective strategy than direct feature merging.

\begin{table}[htbp]
\centering
\caption{Ablation study on Condition-Guided Spatial Transform Fusion (CGSTF) module components on SEVIR dataset. We analyze different fusion strategies and key components within the CGSTF module. The best results are shown in \textbf{bold}.}
\label{tab:cgstf_components4}
\scriptsize  
\setlength{\tabcolsep}{4.5pt}  
\begin{tabular}{l|ccccc}
\hline
\toprule
Method & CSI-M↑ & CSI-181↑ & CSI-219↑ & HSS↑ & MSE↓ \\
\hline
\midrule
w/ Addition & 0.3478 & 0.1889 & 0.0987 & 0.4456 & 441.23 \\
w/ Concatenation & 0.3501 & 0.1923 & 0.1012 & 0.4489 & 439.78 \\
w/o Offset Network & 0.3467 & 0.1876 & 0.0967 & 0.4423 & 443.91 \\
w/o Tanh Activation & 0.3523 & 0.1934 & 0.1045 & 0.4512 & 438.67 \\
\midrule
Full CGSTF & \textbf{0.3552} & \textbf{0.1966} & \textbf{0.1079} & \textbf{0.4576} & \textbf{435.54} \\
\hline
\bottomrule
\end{tabular}
\end{table}

\subsubsection{Evaluation of Wavelet-Guided Skip Connection (WGSC)}
\label{sec:ablation_wgsc}

Table~\ref{tab:wgsc_components5} details the ablation of WGSC components on the CIKM dataset. Removing the wavelet processor—the core of the module—results in a significant performance drop, particularly for high-intensity rainfall (CSI-40 decreases by 7.4\%). This strongly suggests that the frequency-based decomposition into structural (low-frequency) and detail (high-frequency) cues is effective for guiding the reconstruction process. Disabling the condition attention mechanism also leads to a consistent decline in skill, indicating that the gates must be adaptive to the specific conditional input to be fully effective. The largest overall degradation occurs when the final adaptive fusion stage is removed, highlighting the importance of the mechanism that integrates the wavelet cues with the encoder and decoder features. The results confirm that all three components of WGSC—wavelet decomposition, conditional attention, and adaptive fusion—contribute in a complementary manner to its success.

\begin{table}[htbp]
\centering
\caption{Ablation study on WGSC module components on CIKM dataset. We analyze the contribution of key components within the WGSC module. The best results are shown in \textbf{bold}.}
\label{tab:wgsc_components5}
\scriptsize 
\setlength{\tabcolsep}{4.5pt}  %
\begin{tabular}{l|ccccc}
\hline
\toprule
Method & CSI-M↑ & CSI-35↑ & CSI-40↑ & HSS↑ & MSE↓ \\
\hline
\midrule
w/o Wavelet Processor & 0.3198 & 0.2298 & 0.1543 & 0.4156 & 44.67 \\
w/o Condition Attention & 0.3234 & 0.2356 & 0.1598 & 0.4234 & 44.23 \\
w/o Adaptive Fusion & 0.3167 & 0.2267 & 0.1523 & 0.4089 & 44.89 \\
\midrule
Full WGSC & \textbf{0.3293} & \textbf{0.2406} & \textbf{0.1666} & \textbf{0.4278} & \textbf{43.16} \\
\hline
\bottomrule
\end{tabular}
\end{table}

\begin{table}[htbp]
\centering
\scriptsize
\caption{Ablation study on VRWKV deployment strategies on Shanghai dataset. We compare serial and parallel integration schemes with different numbers of VRWKV layers. The best results are shown in \textbf{bold}.}
\label{tab:vrwkv_deployment}
\begin{tabular}{l|ccccc}
\hline
\toprule
Method & CSI-M↑ & CSI-35↑ & CSI-40↑ & HSS↑ & MSE↓ \\
\hline
\midrule
\multicolumn{6}{c}{\textit{Parallel Integration}} \\
\midrule
Parallel 1-layer & 0.4121 & 0.3765 & 0.2736 & 0.5475 & 32.31 \\
Parallel 2-layer & 0.4085 & 0.3716 & 0.2651 & 0.5422 & 31.49 \\
Parallel 3-layer & 0.4115 & 0.3763 & 0.2731 & 0.5483 & 33.01 \\
\hline
\midrule
\multicolumn{6}{c}{\textit{Serial Integration}} \\
\midrule
Serial 1-layer & \textbf{0.4198} & \textbf{0.3867} & \textbf{0.2823} & \textbf{0.5583} & \textbf{32.91} \\
Serial 2-layer & 0.4088 & 0.3713 & 0.2639 & 0.5434 & 31.49 \\
Serial 3-layer & 0.4114 & 0.3728 & 0.2663 & 0.5467 & 32.58 \\
\hline
\multicolumn{6}{c}{\textit{Alternative Modules (serial 1-layer)}} \\
\midrule
Linear attention& 0.4158 & 0.3821 & 0.2790 & 0.5521 & 33.05 \\
SE attention& 0.4137 & 0.3794 & 0.2762 & 0.5510 & 33.12 \\
\hline
\bottomrule
\end{tabular}
\end{table}
\subsubsection{Optimal VRWKV Deployment Strategy} \label{sec:ablation_vrwkv} 
We explored both serial and parallel integration strategies for the VRWKV module at different depths, with results on the Shanghai dataset shown in Table~\ref{tab:vrwkv_deployment}. A serial three-stage deployment—placing one block at each of the encoder tail, the bottleneck, and the first decoder layer—achieves the best overall skill across metrics. This configuration corresponds to the entry denoted Serial 1-layer in the table, where the layer count signifies a single block being applied at each of the three stages. Stacking additional blocks at these same deep stages (i.e., the entries Serial 2-layer and Serial 3-layer) yields diminishing returns, suggesting that once the most compressed representations have been augmented with long-range context, this single-block-per-stage augmentation is sufficient. In contrast, parallel placement is less consistent; for example, two-branch designs tend to lower MSE while degrading high-threshold CSI, indicating smoother forecasts with attenuated intense echoes. To disentangle architectural effects, we also replaced VRWKV with linear attention and with a channel-only SE block at the bottleneck. Both variants produce similar MSE but systematically underperform on CSI/HSS at medium and high thresholds, implying weaker preservation of fine-grained structure. These observations support our default choice: unless otherwise noted, we adopt the serial one-block, three-stage VRWKV configuration.

\section{Conclusion}
\label{sec:conclusion}

This work introduced MFC-RFNet, a radar-only generative architecture for precipitation nowcasting that combines RF training with three complementary design elements: cross-scale communication (FCM), condition-guided shallow alignment (CGSTF), and wavelet-guided skip modulation (WGSC), together with a compact VRWKV block for long-range temporal context. The design aims to coordinate global motion and local evolution while maintaining computational efficiency. Experiments on four public datasets (SEVIR, MeteoNet, Shanghai, CIKM) indicate consistent gains over strong baselines across common metrics. On SEVIR, lead-time curves show slower skill decay at longer horizons, and evaluation at higher thresholds exhibits clear advantages. Component studies further suggest that the improvements arise from complementary effects: CGSTF reduces shallow misalignment, WGSC modulates structure and detail during reconstruction, FCM strengthens cross-scale coherence, and a serial one-block, three-stage VRWKV placement across the encoder tail, the bottleneck, and the first decoder layer offers the best accuracy–efficiency trade-off. The model uses about 27 million parameters and a five-step sampler, providing a practical latency profile.

There are several avenues for future work. First, extending the approach to higher spatial resolutions and longer sequences may require additional memory optimization and multi-scale supervision. Second, robustness under domain shifts (regions, radar products, seasons) could be enhanced through adaptation and calibration strategies. Third, integrating ancillary modalities when available (for example, multisensor radar composites), and exploring alternative frequency analyses or alignment schemes may further improve high-threshold detection. Finally, deploying the model in real-time settings would allow assessment of end-to-end latency and reliability in operational workflows. Overall, the results suggest that coupling RF generation with scale-aware communication, shallow alignment, and frequency-aware fusion is a promising direction for efficient and accurate radar-based nowcasting.
\bibliographystyle{unsrt}

\bibliography{cas-refs}


\end{document}